\documentclass[twoside,11pt]{article}

\usepackage[preprint]{jmlr2e}
\setcitestyle{numbers,square}
\usepackage{amsmath}
\usepackage{amsfonts}
\usepackage{booktabs}
\usepackage{multirow}
\usepackage{enumitem}
\usepackage{xcolor}
\usepackage{microtype}
\usepackage{cleveref}
\usepackage{lastpage}

\hypersetup{
    colorlinks=true,
    linkcolor=blue,
    citecolor=blue,
    urlcolor=blue
}

\newcommand{\adam}{\textsc{Adam}}
\newcommand{\adamw}{\textsc{AdamW}}
\newcommand{\sgd}{\textsc{SGD}}
\newcommand{\adafactor}{\textsc{Adafactor}}
\newcommand{\lion}{\textsc{Lion}}
\newcommand{\sophia}{\textsc{Sophia}}
\newcommand{\muon}{\textsc{Muon}}
\newcommand{\lamb}{\textsc{LAMB}}
\newcommand{\lomo}{\textsc{LOMO}}
\newcommand{\galore}{\textsc{GaLore}}
\newcommand{\R}{\mathbb{R}}
\newcommand{\E}{\mathbb{E}}
\newcommand{\norm}[1]{\left\lVert #1 \right\rVert}

\jmlrheading{0}{2026}{1--\pageref{LastPage}}{Submitted}{Published}{LLM-Valley}{Ranganath}

\ShortHeadings{Navigating LLM Valley}{Ranganath}
\firstpageno{1}

\begin{document}

\title{Navigating LLM Valley: \\
From AdamW to Memory-Efficient and Matrix-Based Optimizers}

\author{\name Aditya Ranganath \email ranganath2@llnl.gov \\
\addr Center for Applied Scientific Computing \\
Lawrence Livermore National Laboratory\\
Livermore, CA, USA}

\editor{Under review}

\maketitle

\begin{abstract}
Training large language models requires optimization algorithms that are not only statistically effective, but also computationally and memory efficient at extreme scale. Although \adamw{} remains the dominant optimizer for large-scale language-model pretraining and fine-tuning, recent work has revisited nearly every component of the optimization stack: adaptive moment estimation, decoupled weight decay, memory footprint, curvature approximation, sign-based updates, large-batch stability, low-rank gradient structure, and matrix-wise orthogonalized updates. This survey reviews optimizer design for large language models through a systems-and-optimization lens. We organize the literature into classical first-order optimizers, adaptive optimizers, memory-efficient variants, second-order and curvature-aware methods, sign-based and discovered optimizers, low-rank and projection-based methods, and matrix-based optimizers such as \muon{}. We also discuss benchmarking methodology, including hyperparameter fairness, scale dependence, wall-clock efficiency, token efficiency, memory overhead, and downstream evaluation. We argue that optimizer research for LLMs is entering a new phase: moving from single-algorithm speedup claims toward rigorous, scale-aware comparisons that jointly evaluate convergence, stability, memory, and implementation complexity.
\end{abstract}



\section{Introduction}
\label{sec:introduction}

Training Large language models (LLMs) requires an optimization process which is \textit{efficiently} scalable to accommodate their size. Modern Transformer-based language models (\cite{brown2020language, chowdhery2023palm, touvron2023llama, touvron2023llama2,dubey2024llama3, jiang2023mistral}) contain billions of parameters, are trained on hundreds of billions or trillions of tokens, and require distributed training systems that combine data parallelism, tensor parallelism, mixed precision and optimizer-state sharding. In this setting, the optimizer is not a minor implementation detail: it directly determines convergence speed, training stability, memory consumption, and ultimately the feasible \textit{learning capability} of a model under a fixed hardware and compute budget.

For most contemporary LLM training pipelines, \adamw{} remains the reference optimizer. Its popularity follows from the empirical success of adaptive moment methods such as \adam{}~\citep{kingma2015adam}, the regularization benefits of decoupled weight decay~\citep{loshchilov2019decoupled}, and its strong performance across Transformer architectures~\citep{vaswani2017attention,devlin2019bert,brown2020language,touvron2023llama}. Compared with plain stochastic gradient descent, adaptive optimizers can better handle heterogeneous gradient statistics across attention projections in a transformer setting (see \cite{zhang2024whytransformersneedadam, tomihari2025gradientheterogeneity}). This robustness has made \adamw{} the default baseline against which most new LLM optimizers are compared.

However, the same features that make \adamw{} effective also make it expensive. Standard \adamw{} maintains first- and second-moment estimates for every trainable parameter. For a model with billions of parameters, these optimizer states can consume memory comparable to or larger than the model weights themselves. The memory burden is particularly acute for full-parameter fine-tuning, long-context training, and pretraining under limited accelerator memory. Even when optimizer states are sharded across devices, the optimizer still affects communication patterns, checkpoint size, numerical precision requirements, and training throughput. Thus, optimizer design for LLMs must be evaluated not only by validation loss, but also by memory footprint, wall-clock time, implementation complexity, and compatibility with distributed systems.

Recent work has therefore reopened the optimizer design space for large-scale language modeling. This work can be distinguished into 4 categories- 1. reduces optimizer-state memory through factorization, quantization, grouping, or fused update rules, as in \adafactor{}~\citep{shazeer2018adafactor}, 8-bit optimizers~\citep{dettmers2022optimizers}, Adam-mini~\citep{zhang2025adammini}, and \lomo{}~\citep{lv2024lomo}, 2. alternatives to conventional coordinate-wise adaptivity, including sign-based and symbolically discovered optimizers such as \lion{}~\citep{chen2023symbolic}, 3. revisits curvature-aware optimization, including scalable second-order or quasi-second-order methods such as Shampoo~\citep{gupta2018shampoo,anil2021scalable} and \sophia{}~\citep{liu2024sophia} and low-rank or matrix structure in Transformer parameters, as in \galore{}~\citep{zhao2024galore} and matrix-based optimizers such as \muon{}~\citep{jordan2024muon,liu2025muon}. Together, these categories suggest that the optimizer for an LLM should not necessarily be viewed as a generic vector update rule; it may instead need to exploit the statistical, architectural, and systems structure of Transformer training.

Despite this growing literature, optimizer research for LLMs remains fragmented. General optimization surveys provide broad background on stochastic, adaptive, and second-order methods~\citep{ruder2016overview,bottou2018optimization,sun2019survey}, while Transformer-efficiency surveys discuss optimizers as one component among many training-acceleration techniques~\citep{tay2023efficient,wan2023efficienttraining}. However, LLM optimizer papers often differ in model scale, token budget, architecture, dataset, batch size, learning-rate schedule, precision, hardware setup, and hyperparameter tuning effort. These differences make it difficult to determine whether a reported gain reflects a fundamentally better optimizer, a better-tuned baseline, a favorable small-scale regime, or a systems-level implementation advantage. Recent benchmarking studies have emphasized that optimizer comparisons can be highly sensitive to tuning and scale, and that purported gains may shrink under fairer or larger-scale evaluation~\citep{zhao2025deconstructing,semenov2025benchmarking}.

This survey focuses specifically on \emph{weight-update optimizers} for LLM pretraining and full-parameter fine-tuning. We use the term optimizer to mean an algorithm that updates trainable model parameters from gradients or gradient-derived statistics. This scope includes classical first-order methods, adaptive diagonal methods, large-batch optimizers, memory-efficient variants, sign-based and discovered update rules, curvature-aware methods, low-rank projection methods, and matrix-based optimizers. We do not attempt to comprehensively survey parameter-efficient fine-tuning, pruning, quantization for inference, or decoding-time acceleration, except where such methods directly interact with optimizer design or optimizer-state memory.

The goal of this survey is to organize the emerging optimizer landscape for LLMs into a coherent taxonomy. In particular, we make the following contributions:
\begin{enumerate}[leftmargin=*]
    \item We review the role of optimizers in LLM training and identify the LLM-specific constraints that distinguish this setting from conventional small-scale neural-network optimization.
    \item We categorize LLM optimizers by update geometry, memory cost, curvature usage, scale behavior, and exploitation of matrix or low-rank structure.
    \item We compare the motivations and trade-offs of representative methods, including \adamw{}, \adafactor{}, \lamb{}, \lion{}, \sophia{}, \lomo{}, \galore{}, Adam-mini, and \muon{}.
    \item We discuss methodological challenges in optimizer benchmarking, including hyperparameter fairness, scale dependence, token efficiency versus wall-clock efficiency, memory accounting, and downstream evaluation.
    \item We identify open problems for future optimizer research, including optimizer scaling laws, parameter-group-specific update rules, matrix-aware theory, and system-level comparisons under fixed compute and memory budgets.
\end{enumerate}

We organize the remainder of the survey as follows. \Cref{sec:background} introduces notation and reviews the optimization constraints that arise in LLM training. \Cref{sec:taxonomy} presents our taxonomy of optimizer families. \Cref{sec:adaptive} discusses classical and adaptive first-order optimizers, including \sgd{}, \adam{}, \adamw{}, and \adafactor{}. \Cref{sec:largebatch} covers large-batch and distributed-training optimizers. \Cref{sec:memory} reviews memory-efficient optimizers. \Cref{sec:sign} discusses sign-based and discovered update rules. \Cref{sec:curvature} examines curvature-aware and second-order methods. \Cref{sec:lowrank} covers low-rank and projection-based optimizers. \Cref{sec:matrix} discusses matrix-based and orthogonalized optimizers. \Cref{sec:benchmarking} analyzes evaluation methodology, and \Cref{sec:openproblems} concludes with open research directions.

\section{Background: Optimization in LLM Training}
\label{sec:background}

This section introduces the optimization setting for large language model training. We first formalize next-token prediction as stochastic optimization, then review the update rules underlying common first-order optimizers. We then discuss why LLM training creates distinctive constraints for optimizer design, including optimizer-state memory, mixed precision, distributed training, parameter heterogeneity, and long-horizon stability.

\subsection{Language-Model Training as Stochastic Optimization}
\label{subsec:lm_stochastic_optimization}

Let $\theta \in \R^d$ denote the trainable parameters of a language model. Given a token sequence $x_{1:T}$, an autoregressive language model factorizes the probability of the sequence as
\begin{equation}
    p_\theta(x_{1:T}) = \prod_{t=1}^{T} p_\theta(x_t \mid x_{<t}).
\end{equation}
Training typically minimizes the empirical next-token negative log-likelihood,
\begin{equation}
    \mathcal{L}(\theta)
    = - \E_{x_{1:T} \sim \mathcal{D}} \left[ \sum_{t=1}^{T} \log p_\theta(x_t \mid x_{<t}) \right],
\end{equation}
where $\mathcal{D}$ is the training distribution. In practice, training uses minibatches of token sequences, producing a stochastic loss $\mathcal{L}_t(\theta)$ and stochastic gradient
\begin{equation}
    g_t = \nabla_\theta \mathcal{L}_t(\theta_t).
\end{equation}
An optimizer defines an update map
\begin{equation}
    \theta_{t+1} = \mathcal{U}(\theta_t, g_t, s_t; \eta_t, \lambda, \phi),
\end{equation}
where $s_t$ denotes optimizer state, $\eta_t$ is the learning rate, $\lambda$ is a weight-decay coefficient, and $\phi$ denotes additional optimizer hyperparameters such as momentum coefficients, numerical stabilizers, clipping thresholds, or preconditioner update frequencies.

For LLMs, the optimizer is typically coupled to a learning-rate schedule. A common schedule includes a warmup phase followed by cosine decay or inverse-square-root decay. Warmup is especially important for adaptive optimizers because moment estimates are initially inaccurate and because large Transformer models can be numerically unstable early in training.

\subsection{Stochastic Gradient Descent and Momentum}
\label{subsec:bg_sgd_momentum}

The simplest optimizer is stochastic gradient descent:
\begin{equation}
    \theta_{t+1} = \theta_t - \eta_t g_t.
\end{equation}
Although \sgd{} is memory efficient, it uses a single scalar learning rate for all coordinates and does not adapt to heterogeneous gradient scales. This limitation is significant in LLMs, where different parameter groups can exhibit substantially different gradient statistics.

Momentum improves \sgd{} by accumulating a velocity vector:
\begin{align}
    v_t &= \beta v_{t-1} + g_t, \\
    \theta_{t+1} &= \theta_t - \eta_t v_t,
\end{align}
where $\beta \in [0,1)$ controls the decay of previous gradients. Momentum can reduce oscillation and improve optimization in poorly conditioned landscapes. However, it still does not provide coordinate-wise adaptivity.

\subsection{Adaptive Diagonal Preconditioning}
\label{subsec:bg_adaptive_preconditioning}

Adaptive optimizers can be interpreted as diagonal preconditioned gradient methods. Instead of applying the same learning rate to every coordinate, they maintain statistics of past gradients and rescale updates coordinate-wise.

AdaGrad accumulates squared gradients,
\begin{align}
    G_t &= G_{t-1} + g_t^2, \\
    \theta_{t+1} &= \theta_t - \eta_t \frac{g_t}{\sqrt{G_t} + \epsilon},
\end{align}
where operations are element-wise~\citep{duchi2011adagrad}. RMSProp replaces the cumulative sum with an exponential moving average~\citep{tieleman2012rmsprop},
\begin{equation}
    v_t = \beta_2 v_{t-1} + (1-\beta_2) g_t^2.
\end{equation}

\adam{} combines momentum and adaptive second-moment scaling~\citep{kingma2015adam}. It maintains first- and second-moment estimates,
\begin{align}
    m_t &= \beta_1 m_{t-1} + (1 - \beta_1) g_t, \\
    v_t &= \beta_2 v_{t-1} + (1 - \beta_2) g_t^2,
\end{align}
with bias-corrected moments
\begin{align}
    \hat{m}_t &= \frac{m_t}{1 - \beta_1^t}, \\
    \hat{v}_t &= \frac{v_t}{1 - \beta_2^t}.
\end{align}
The parameter update is
\begin{equation}
    \theta_{t+1} = \theta_t - \eta_t \frac{\hat{m}_t}{\sqrt{\hat{v}_t} + \epsilon}.
\end{equation}
The second-moment accumulator $v_t$ acts as a diagonal estimate of gradient scale and yields coordinate-wise learning-rate adaptation.

\subsection{AdamW and Decoupled Weight Decay}
\label{subsec:bg_adamw}

Weight decay is often used to regularize neural-network training. In classical \sgd{}, $\ell_2$ regularization and multiplicative weight decay are closely related. In adaptive optimizers, however, coupling weight decay to the gradient update changes the effective regularization because the penalty term is also rescaled by the adaptive preconditioner. \adamw{} addresses this issue by decoupling weight decay from the adaptive gradient update~\citep{loshchilov2019decoupled}.

A simplified \adamw{} update can be written as
\begin{equation}
    \theta_{t+1}
    = (1 - \eta_t \lambda)\theta_t
    - \eta_t \frac{\hat{m}_t}{\sqrt{\hat{v}_t} + \epsilon},
\end{equation}
where $\lambda$ is the decoupled weight-decay coefficient. This separation allows the adaptive update and regularization strength to be tuned more independently. In LLM training, \adamw{} is commonly combined with learning-rate warmup, cosine decay, gradient clipping, parameter-specific weight-decay exclusions, and mixed-precision arithmetic.

\subsection{Optimizer-State Memory}
\label{subsec:bg_optimizer_memory}

The memory cost of an optimizer is a crucial issue in LLM training. Suppose a model has $d$ parameters. Ignoring activation memory, a simple mixed-precision training setup may store model parameters, gradients, master weights, and optimizer states. For \adamw{}, the optimizer state includes at least two vectors of length $d$: the first moment $m_t$ and the second moment $v_t$.

A simplified memory accounting is shown in \Cref{tab:optimizer_memory_background}. The exact memory footprint depends on precision, framework implementation, parameter sharding, and whether master weights are stored separately. Nevertheless, the qualitative point is robust: adaptive optimizers can require multiple additional copies of the parameter vector.

\begin{table}[t]
\centering
\small
\begin{tabular}{lcc}
\toprule
\textbf{Stored quantity} & \textbf{Typical precision} & \textbf{Approximate cost} \\
\midrule
Model parameters & bf16/fp16 & $2d$ bytes \\
Gradients & bf16/fp16 & $2d$ bytes \\
Master parameters & fp32 & $4d$ bytes \\
First moment $m_t$ & fp32 & $4d$ bytes \\
Second moment $v_t$ & fp32 & $4d$ bytes \\
\midrule
Total, unsharded AdamW-style training & -- & up to $16d$ bytes \\
\bottomrule
\end{tabular}
\caption{Simplified memory accounting for mixed-precision \adamw{}-style training. Actual implementations may differ, but first- and second-moment states are a major source of memory overhead.}
\label{tab:optimizer_memory_background}
\end{table}

For a billion-parameter model, each fp32 optimizer-state vector requires approximately 4 GB of memory. Thus, the two moment vectors alone require approximately 8 GB before accounting for parameters, gradients, activations, fragmentation, and distributed-training overhead. This scaling motivates memory-efficient optimizers such as \adafactor{}~\citep{shazeer2018adafactor}, low-bit optimizers~\citep{dettmers2022optimizers}, Adam-mini~\citep{zhang2025adammini}, and full-parameter fine-tuning methods such as \lomo{}~\citep{lv2024lomo}.

\subsection{Mixed Precision and Numerical Stability}
\label{subsec:bg_mixed_precision}

LLMs are commonly trained using mixed precision, such as fp16, bf16, or fp8 variants, to reduce the memory footprint and increase throughput. However, this changes the numerical behavior of optimization. 

Gradients may underflow or overflow, moment estimates may lose precision, and update magnitudes may become unstable if the optimizer is not implemented carefully.

In mixed-precision training, gradients may underflow or overflow because fp16 has a limited numerical range; standard remedies include loss scaling and maintaining fp32 master weights~\citep{micikevicius2018mixed,zhao2019adaptive}. Optimizer states are also precision-sensitive: low-bit Adam-style methods require careful quantization to preserve moment statistics and training stability~\citep{dettmers2022optimizers}. Finally, Adam-style update magnitudes can become unstable at large scale or under low-precision arithmetic if numerical details such as epsilon, clipping, and state precision are not handled carefully~\citep{molybog2023adaminstability,wang2023stableadam16bit}.

Adaptive optimizers introduce additional numerical considerations. The stabilizer $\epsilon$ in \adam{} and \adamw{} can meaningfully affect training when second-moment estimates are small. Gradient clipping is often used to prevent rare but large updates. Loss scaling is commonly used with fp16 training, while bf16 often reduces the need for explicit scaling because of its larger exponent range. Optimizer comparisons should therefore specify precision, clipping rules, loss scaling, and the exact implementation of moment updates.

\subsection{Distributed Training and Optimizer Sharding}
\label{subsec:bg_distributed_training}

Large LLMs often require modern GPUs and servers which accelerate the training process significantly. With a huge model footprint, these models require sophisticated data and model parallelism. This data and model parallelism also extend to the optimizer states. 

\medskip

\noindent \textbf{Gradient synchronization.} In data-parallel LLM training, each worker computes gradients on a local microbatch, and these gradients must be synchronized before the optimizer step. The standard approach is an all-reduce operation (\cite{sergeev2018horovod}) that averages gradients across workers, ensuring that every replica applies the same update. Gradient synchronization can become a major communication bottleneck as model size and worker count increase, especially when gradients are communicated in full precision or when synchronization occurs after every backward pass. Practical systems reduce this cost through gradient accumulation, overlapping communication with backpropagation, bucketed all-reduce, mixed-precision communication, and optimizer-state sharding. Optimizer design interacts with this process because methods with additional statistics, structured pre-conditioners, or matrix-level updates may require extra synchronization beyond ordinary gradient averaging (\cite{pytorchddp,shoeybi2019megatron, rajbhandari2020zero}).

\medskip 

\noindent \textbf{Model sharding.} Model sharding partitions model-related tensors across devices so that training is no longer limited by the memory capacity of a single accelerator. In contrast to ordinary data parallelism, which replicates parameters, gradients, and optimizer states on every worker, sharded training distributes some or all of these tensors across workers and reconstructs them only when needed for computation. ZeRO introduced a staged approach to remove redundancy in optimizer states, gradients, and parameters, enabling much larger models while preserving data-parallel training semantics~\citep{rajbhandari2020zero}. PyTorch Fully Sharded Data Parallel (FSDP) brings similar ideas into a native PyTorch implementation, sharding parameters, gradients, and optimizer states while integrating with autograd, CUDA memory management, prefetching, and communication--computation overlap~\citep{zhao2023fsdp}. In parallel, Megatron-style tensor and pipeline parallelism shard individual layers and sequences across devices, making it possible to train multi-billion- and trillion-parameter Transformer models by combining data, tensor, pipeline, and sequence parallelism~\citep{shoeybi2019megatron,narayanan2021efficient}. Recent development has moved toward hybrid sharding strategies that combine FSDP/ZeRO-style data-parallel sharding with tensor, pipeline, context, and sequence parallelism. Thus, optimizer designs must now be evaluated together with the distributed layout that stores and synchronizes its states.

\noindent \textbf{Optimizer sharding.} Optimizer sharding partitions the auxiliary state used by the optimizer, such as Adam first- and second-moment tensors, across data-parallel workers. This differs from model sharding, which partitions the model parameters themselves, often through tensor, pipeline, or fully sharded data parallelism. In ordinary data parallelism, every worker stores a full copy of the parameters, gradients, and optimizer states; optimizer sharding removes redundancy in the optimizer states while preserving the semantics of data-parallel training. ZeRO formalizes this distinction through staged partitioning: ZeRO-1 shards optimizer states, ZeRO-2 additionally shards gradients, and ZeRO-3 shards parameters as well~\citep{rajbhandari2020zero}. In practice, optimizer sharding is especially important for Adam-style optimizers because moment estimates can consume memory comparable to or larger than the model weights, while model sharding is needed when the parameters themselves no longer fit on a single accelerator~\citep{zhao2023fsdp,narayanan2021efficient}.

Even though gradient synchronization and model sharding reduce per-device memory and compute, they introduce communication and implementation complexity. Thus, the cost of an optimizer is not fully captured by its mathematical update rule. An optimizer with fewer states may enable larger batch sizes or longer context lengths. An optimizer with matrix preconditioning may reduce optimization steps but introduce additional communication or computation. A fair comparison must include not only validation loss per token, but also wall-clock time, peak memory, communication volume, checkpoint size, and compatibility with existing distributed-training infrastructure.

\subsection{Parameter Heterogeneity in Transformers}
\label{subsec:bg_parameter_heterogeneity}

Transformer LLMs contain parameter groups with different shapes and roles. Some of these include token embeddings, positional embedding parameters, projection matrices, and output heads. Standard optimizers usually apply a shared update rule across all parameters with simple exclusions for weight decay. However, different parameter groups may have different gradient scales, curvature, rank structure, and sensitivity to regularization.

This heterogeneity motivates several recent optimizer directions. Factorized optimizers exploit matrix-shaped parameters. Low-rank optimizers exploit gradient structure in large dense matrices. Matrix-based optimizers such as \muon{} treat certain weight tensors as matrices and modify the geometry of their updates~\citep{jordan2024muon,liu2025muon}. Parameter grouping in Adam-mini similarly reflects the possibility that different parts of the model may not require fully independent coordinate-wise adaptivity~\citep{zhang2025adammini}.

\subsection{Token Efficiency, Wall-Clock Efficiency, and Final Performance}
\label{subsec:bg_efficiency_metrics}

An optimizer for LLMs can be evaluated along multiple axes. Token efficiency measures how quickly validation loss decreases as a function of the number of training tokens. Step efficiency measures progress per optimizer step. Wall-clock efficiency measures progress per unit time. Memory efficiency measures the largest model, context length, or batch size that can fit under a fixed memory budget. Final performance measures downstream quality after a fixed training recipe.

These metrics can disagree. A curvature-aware optimizer may improve token efficiency but slow each step. A memory-efficient optimizer may allow larger batches or models but require more tokens to reach the same loss. A new optimizer may outperform \adamw{} at small scale but lose its advantage when both methods are carefully tuned at larger scale. Recent empirical studies emphasize the importance of fair hyperparameter tuning, scale-aware evaluation, and cautious interpretation of early training curves~\citep{zhao2025deconstructing,semenov2025benchmarking}.

\subsection{Summary}
\label{subsec:bg_summary}

LLM optimization is shaped by both algorithmic and systems constraints. The optimizer must produce stable and effective parameter updates, but it must also fit within strict memory budgets, interact well with mixed precision, scale across distributed hardware, and remain robust across heterogeneous Transformer parameter groups. These considerations explain why \adamw{} remains a strong baseline, and why recent work has explored memory-efficient, curvature-aware, low-rank, and matrix-based alternatives. The next section uses these dimensions to construct a taxonomy of LLM optimizers.

\section{A Taxonomy of Optimizers for LLMs}
\label{sec:taxonomy}

The optimizer landscape for LLMs is too diverse to be organized by chronology alone. Many recent methods modify different parts of the optimization pipeline: some change the update direction, some change the preconditioner, some reduce optimizer-state memory, some exploit matrix structure, and others target distributed training or full-parameter fine-tuning under memory constraints. This section introduces a taxonomy that will guide the rest of the survey.

The LLM optimizers are organized along five axes:
\begin{enumerate}[leftmargin=*]
    \item \textbf{Update geometry:} whether the optimizer uses raw gradients, momentum, coordinate-wise adaptive scaling, sign updates, curvature approximations, low-rank projections, or matrix-structured transformations.
    \item \textbf{State memory:} how much auxiliary state the optimizer stores relative to the number of trainable parameters.
    \item \textbf{Structural assumptions:} whether the method treats parameters as independent coordinates, vectors, matrices, tensors, low-rank objects, or parameter groups.
    \item \textbf{Target regime:} whether the optimizer is intended primarily for pretraining, full-parameter fine-tuning, large-batch training, memory-constrained training, or general-purpose use.
    \item \textbf{Evaluation criterion:} whether the method aims to improve token efficiency, wall-clock speed, peak memory, numerical stability, final downstream performance, or implementation simplicity.
\end{enumerate}

These axes are not independent. For example, a method that reduces memory by compressing optimizer states may change numerical behavior; a curvature-aware optimizer may improve token efficiency but increase per-step compute; and a matrix-based optimizer may exploit Transformer weight structure while requiring specialized kernels or update rules. Therefore, evaluating the best optimizer based on its lowest validation loss might be of significance given many of the other factors taken into consideration. 

\subsection{Classical First-Order Methods}
\label{subsec:taxonomy_classical}

Classical first-order methods update parameters directly from stochastic gradients, optionally with momentum. This family includes \sgd{}, momentum, and Nesterov-style acceleration. Their main advantage is simplicity and low memory overhead: plain \sgd{} requires no per-parameter auxiliary state, and momentum requires only one additional vector. Their main limitation for LLMs is the lack of coordinate-wise adaptivity. Transformer parameters often have heterogeneous gradient scales across layers and parameter types, making a single scalar learning rate difficult to tune.

Although \sgd{} remains important as a conceptual and theoretical baseline, it is rarely the default choice for modern LLM pretraining. Recent empirical studies often find that \sgd{} performs worse than adaptive methods for autoregressive language modeling when compared under practical training recipes~\citep{zhao2025deconstructing}. Nevertheless, \sgd{}-like memory usage remains attractive, motivating methods that attempt to approach \adamw{}-level performance with reduced optimizer state.

\subsection{Adaptive Diagonal Methods}
\label{subsec:taxonomy_adaptive}

Adaptive diagonal optimizers maintain coordinate-wise statistics of past gradients and use them to rescale updates. This family includes AdaGrad~\citep{duchi2011adagrad}, RMSProp~\citep{tieleman2012rmsprop}, \adam{}~\citep{kingma2015adam}, \adamw{}~\citep{loshchilov2019decoupled}, and \adafactor{}~\citep{shazeer2018adafactor}. The dominant member of this family in LLM training is \adamw{}.

The strength of adaptive diagonal methods is robustness. By normalizing coordinates according to gradient magnitude, they can tolerate heterogeneous parameter scales and sparse or bursty gradients. Their weakness is memory: \adamw{} typically stores two full-size moment vectors, which is costly for billion-parameter models. Much of the recent optimizer literature can be understood as an attempt to preserve the benefits of adaptive diagonal scaling while reducing or restructuring its memory cost.

\subsection{Large-Batch and Distributed-Training Optimizers}
\label{subsec:taxonomy_largebatch}

Large-batch optimizers aim to maintain stable optimization when increasing batch size to improve hardware utilization. This family includes methods such as \lamb{}, which introduces layer-wise trust ratios and was used to train BERT with very large batches~\citep{you2020large}. These optimizers are motivated by the observation that naive learning-rate scaling can become unstable as batch size grows.

For LLMs, large-batch optimization is intertwined with distributed systems. The relevant question is not only whether an optimizer converges with a large batch, but whether it improves throughput under data parallelism, tensor parallelism, pipeline parallelism, and optimizer-state sharding. Large-batch optimizers are therefore best understood as system-aware optimization methods.

\subsection{Memory-Efficient Optimizers}
\label{subsec:taxonomy_memory}

Memory-efficient optimizers reduce the cost of optimizer states, gradients, or update computation. This family includes factorized-state methods such as \adafactor{}~\citep{shazeer2018adafactor}, quantized-state methods such as 8-bit optimizers~\citep{dettmers2022optimizers}, reduced-adaptivity methods such as Adam-mini~\citep{zhang2025adammini}, and fused full-parameter fine-tuning methods such as \lomo{}~\citep{lv2024lomo}.

The motivation is straightforward: optimizer state can determine whether a model fits in memory. Reducing optimizer memory may enable larger models, longer contexts, larger batches, or full-parameter fine-tuning on limited hardware. The trade-off is that memory reduction may also reduce update fidelity, require special parameter grouping, introduce quantization error, or complicate implementation.

\subsection{Sign-Based and Discovered Optimizers}
\label{subsec:taxonomy_sign}

Sign-based optimizers use the sign of a gradient or momentum-like quantity rather than its full magnitude. Examples include signSGD~\citep{bernstein2018signsgd} and \lion{}, which was discovered through symbolic search over optimizer programs~\citep{chen2023symbolic}. These optimizers are attractive because sign operations are simple and can reduce dependence on precise gradient magnitudes.

In LLM training, sign-based methods raise two questions - 1. Can they match the robustness of \adamw{} across model scales and architectures? 2. do their apparent gains persist under fair hyperparameter tuning and large-scale evaluation? The importance of these questions may be attributed to the optimizer's sensitivity to learning rate, weight decay, and schedule choices.

\subsection{Curvature-Aware and Second-Order Methods}
\label{subsec:taxonomy_curvature}

Curvature-aware optimizers attempt to improve update directions by approximating second-order information. Exact Newton-style optimization is infeasible for LLMs because the Hessian is enormous. If we revisit our one billion parameter model in \ref{subsec:bg_optimizer_memory} - for a model with \(d=10^9\) parameters, a dense Hessian would contain \(10^{18}\) entries; even storing these entries in fp32 would require roughly \(4 \times 10^{18}\) bytes, i.e., exabytes of memory. Exact Hessian storage, communication, and inversion are therefore completely infeasible for LLMs. Practical methods therefore use diagonal, block-diagonal, Kronecker-factored, or tensor-structured approximations.

This family includes diagonal curvature methods such as Sophia, tensor-structured preconditioners such as Shampoo, and SOAP-like structured methods. Quasi-Newton methods are closely related, but we treat them as a separate subfamily in Section \ref{sec:taxonomy} because they estimate inverse-curvature information from parameter and gradient differences rather than directly accumulating Hessian, Fisher, or gradient-covariance statistics.

\subsection{Low-Rank and Projection-Based Optimizers}
\label{subsec:taxonomy_lowrank}

Low-rank and projection-based methods exploit the empirical observation that gradients or updates for large neural-network matrices may have low effective rank. \galore{}, for example, projects gradients into a low-rank subspace to reduce memory while maintaining full-parameter training~\citep{zhao2024galore}. These methods occupy an intermediate position between optimizer design and memory-efficient training.

The key design question is how to choose and update the projection subspace. A projection that is too small may lose important update information, while a projection that is too large may provide limited memory savings. These methods are particularly relevant for Transformer layers because attention and feed-forward blocks contain large dense matrices where low-rank structure may be exploitable.

\subsection{Matrix-Based and Orthogonalized Optimizers}
\label{subsec:taxonomy_matrix}

Matrix-based optimizers treat weight matrices as matrices rather than collections of independent scalar parameters. This distinguishes them from coordinate-wise adaptive optimizers, which apply element-wise transformations. \muon{} is a representative recent method that applies orthogonalization or matrix-normalization ideas to momentum updates for neural-network weight matrices~\citep{jordan2024muon,liu2025muon}.

This family is motivated by the structure of Transformer architectures. Most parameters in LLMs reside in large matrices: query, key, value, output, and feed-forward projections. Matrix-based optimizers ask whether update geometry should respect this structure. Their potential advantage is better-conditioned updates for matrix parameters. Their challenge is to demonstrate robust gains at LLM scale while keeping computational overhead and implementation complexity manageable.

\subsection{Quasi-Newton Optimizers}
\label{subsec:taxonomy_quasi_newton}

Quasi-Newton optimizers approximate second-order information without explicitly forming the Hessian. Classical methods such as BFGS construct an approximation to the inverse Hessian from successive parameter differences and gradient differences, while limited-memory variants such as L-BFGS store only a small history of such pairs. This makes quasi-Newton methods conceptually attractive for large models because they can improve conditioning without storing a full curvature matrix\citep{liu1989lbfgs,nocedal2006numerical,schraudolph2007stochastic,mokhtari2015global,goldfarb2020practical,niu2023mlbfgs,ranganath2025sr1cubic}.

For LLMs, however, quasi-Newton methods face several challenges. Stochastic minibatch gradients can make curvature-pair estimates noisy, line-search procedures are difficult to integrate into large-scale distributed training, and even limited-memory histories can be expensive at billion-parameter scale. Recent work on stochastic and distributed quasi-Newton methods attempts to address these issues through block-wise approximations, momentum stabilization, and distributed memory/computation strategies. Thus, quasi-Newton optimizers occupy an intermediate position between first-order adaptive methods and explicit curvature-aware preconditioners: they use curvature information implicitly through update histories, but must be adapted carefully for stochastic, large-scale, distributed LLM training.

\subsection{Comparison of Optimizer Families}
\label{subsec:taxonomy_comparison}

\Cref{tab:optimizer_taxonomy} summarizes the main optimizer families discussed in this survey. The table emphasizes that optimizer families differ not only in mathematical update rules, but also in memory cost, structural assumptions, and target use cases.

\begin{table}[t]
\centering
\small
\begin{tabular}{p{0.18\linewidth}p{0.20\linewidth}p{0.20\linewidth}p{0.30\linewidth}}
\toprule
\textbf{Family} & \textbf{Representative methods} & \textbf{State cost} & \textbf{Primary motivation} \\
\midrule
Classical first-order & \sgd{}, Momentum, Nesterov & None or one vector & Simple, low-memory baselines. \\
Adaptive diagonal & AdaGrad, RMSProp, \adam{}, \adamw{}, \adafactor{} & One to two vectors; factorized for \adafactor{} & Robust coordinate-wise scaling for heterogeneous gradients. \\
Large-batch & \lamb{}, LARS-style methods & Usually adaptive-state dependent & Stable training at large batch sizes and high hardware utilization. \\
Memory-efficient & \adafactor{}, 8-bit Adam, Adam-mini, \lomo{} & Reduced, compressed, grouped, or fused states & Fit larger models, longer contexts, or full fine-tuning under memory limits. \\
Sign-based/discovered & signSGD, \lion{} & Typically one or two vectors & Simpler update rules and possible speed or memory benefits. \\
Curvature-aware & Shampoo, SOAP-like methods, \sophia{} & Diagonal, block, or structured curvature states & Better conditioning and improved token efficiency. \\
Low-rank/projection & \galore{}, APOLLO-like methods & Low-rank subspace states & Reduce training memory by exploiting low-rank gradient structure. \\
Matrix-based & \muon{} and related methods & Matrix-specific state or transformations & Exploit weight-matrix geometry and orthogonalized updates. \\
Quasi-Newton methods & BFGS, L-BFGS, stochastic L-BFGS, mL-BFGS & Limited history of parameter/gradient differences & Approximate inverse-Hessian information without full second-order storage. \\
\bottomrule
\end{tabular}
\caption{Taxonomy of optimizer families for LLM training and fine-tuning. State cost is qualitative because actual memory depends on precision, sharding, parameter grouping, and implementation.}
\label{tab:optimizer_taxonomy}
\end{table}

\subsection{A Systems-Oriented View}
\label{subsec:taxonomy_systems_view}

A useful way to compare optimizers is to view them as points in a systems--optimization design space. At one extreme, \sgd{} is simple and memory efficient but often less robust for LLM training. At another extreme, second-order and structured-preconditioning methods may provide better-conditioned updates but add computational and implementation overhead. \adamw{} occupies a strong middle ground: it is robust and well supported, but memory intensive. Many recent optimizers attempt to move away from \adamw{} along one of three directions: reducing memory while preserving adaptive behavior, improving update geometry while controlling overhead, or exploiting architecture-specific structure.

This perspective also clarifies why no single metric is sufficient. Validation loss at a fixed number of tokens measures token efficiency, but not wall-clock efficiency. Wall-clock time measures practical speed, but depends on implementation quality and hardware. Peak memory determines feasible model scale, but may trade off against convergence. Downstream performance measures utility, but can hide training instability or cost. A rigorous comparison must therefore report several metrics simultaneously.

\subsection{Summary}
\label{subsec:taxonomy_summary}

The optimizer landscape for LLMs can be organized by update geometry, memory cost, structural assumptions, target training regime, and evaluation criterion. This taxonomy frames the rest of the survey. The next section begins with classical and adaptive first-order methods, focusing on why \adamw{} became the dominant baseline and how alternatives such as \adafactor{} modify the adaptive-optimization template.
\section{Classical and Adaptive First-Order Optimizers}
\label{sec:adaptive}

Classical and adaptive first-order optimizers form the foundation of modern LLM training. Even newer optimizer families---including memory-efficient, low-rank, curvature-aware, and matrix-based methods---are often best understood as modifications of first-order updates. This section reviews the core methods in this family, explains why \adamw{} became the dominant optimizer for LLMs, and discusses the role of \adafactor{} as an important memory-reduced adaptive alternative.

\subsection{SGD as the Baseline Update Rule}
\label{subsec:sgd}

Stochastic gradient descent updates parameters in the negative direction of the minibatch gradient:
\begin{equation}
    \theta_{t+1} = \theta_t - \eta_t g_t,
\end{equation}
where $g_t = \nabla_\theta \mathcal{L}_t(\theta_t)$ and $\eta_t$ is the learning rate. The appeal of \sgd{} is its simplicity. It requires no auxiliary optimizer state, has low memory overhead, and is easy to implement in distributed training systems. In terms of memory, plain \sgd{} is ideal: aside from parameters and gradients, it need not store moment estimates, curvature approximations, or projection matrices.

However, this simplicity is also a limitation. \sgd{} applies the same scalar learning rate to all parameters. This can be problematic in LLMs, where parameter groups differ dramatically in scale, frequency of update, gradient variance, and curvature. Token embeddings, attention projections, feed-forward matrices, normalization parameters, and output heads may all require different effective step sizes. A single global learning rate must therefore compromise across heterogeneous parameter groups.

Momentum partially addresses this problem by accumulating a velocity vector,
\begin{align}
    u_t &= \beta u_{t-1} + g_t, \\
    \theta_{t+1} &= \theta_t - \eta_t u_t,
\end{align}
where $\beta$ is the momentum coefficient. Momentum reduces high-frequency noise in the update direction and can accelerate progress along persistent descent directions. Nevertheless, it still lacks coordinate-wise adaptivity. It stores one additional vector, making it more memory intensive than plain \sgd{} but still less expensive than \adamw{}.

In LLM pretraining, \sgd{} and momentum are useful as conceptual baselines and for understanding the value of adaptivity. Empirical comparisons of optimizers for autoregressive language modeling generally find that \sgd{} is difficult to make competitive with adaptive methods under practical training recipes~\citep{zhao2025deconstructing}. This observation motivates a recurring theme in LLM optimizer research: the goal is often not to return to plain \sgd{}, but to obtain \sgd{}-like memory cost while preserving some of the robustness of adaptive methods.

\subsection{Coordinate-Wise Adaptivity}
\label{subsec:coordinate_adaptivity}

Adaptive optimizers maintain per-coordinate statistics of gradients and use those statistics to rescale updates. The generic form is
\begin{equation}
    \theta_{t+1} = \theta_t - \eta_t D_t^{-1} h_t,
\end{equation}
where $h_t$ is a gradient-derived direction, often a momentum estimate, and $D_t$ is a diagonal preconditioner derived from past gradients. In this view, adaptivity changes the geometry of the update: coordinates with historically large gradients receive smaller effective learning rates, while coordinates with small gradients receive larger effective learning rates.

AdaGrad accumulates squared gradients over time~\citep{duchi2011adagrad}:
\begin{align}
    r_t &= r_{t-1} + g_t^2, \\
    \theta_{t+1} &= \theta_t - \eta_t \frac{g_t}{\sqrt{r_t} + \epsilon}.
\end{align}
This update can be effective for sparse features, but the accumulator $r_t$ grows monotonically, causing effective learning rates to decay continually. RMSProp replaces the cumulative sum with an exponential moving average~\citep{tieleman2012rmsprop}:
\begin{equation}
    r_t = \beta_2 r_{t-1} + (1 - \beta_2) g_t^2.
\end{equation}
This makes the preconditioner responsive to recent gradient statistics rather than the full training history.

The key advantage of coordinate-wise adaptivity is robustness to heterogeneous gradient scales. This is especially valuable for Transformers, whose parameters play different functional roles and may experience different gradient distributions. The main cost is memory: a diagonal preconditioner requires storing at least one auxiliary value per parameter. Adaptive methods also introduce additional hyperparameters, such as $\beta_2$ and $\epsilon$, whose effects can be nontrivial in mixed-precision training.

\subsection{Adam}
\label{subsec:adam}

\adam{} combines momentum and adaptive second-moment scaling~\citep{kingma2015adam}. It maintains exponential moving averages of gradients and squared gradients:
\begin{align}
    m_t &= \beta_1 m_{t-1} + (1 - \beta_1)g_t, \\
    v_t &= \beta_2 v_{t-1} + (1 - \beta_2)g_t^2.
\end{align}
The bias-corrected estimates are
\begin{align}
    \hat{m}_t &= \frac{m_t}{1 - \beta_1^t}, \\
    \hat{v}_t &= \frac{v_t}{1 - \beta_2^t}.
\end{align}
The update is
\begin{equation}
    \theta_{t+1}
    = \theta_t - \eta_t \frac{\hat{m}_t}{\sqrt{\hat{v}_t}+\epsilon}.
\end{equation}

The first-moment estimate $m_t$ smooths stochastic gradients, while the second-moment estimate $v_t$ rescales coordinates. This combination has made \adam{} effective across a broad range of neural-network architectures. However, \adam{} also raised theoretical and practical questions. In particular, adaptive methods can fail to converge under some settings, motivating variants such as AMSGrad~\citep{reddi2018adam}. In LLM practice, the more influential modification is not AMSGrad but decoupled weight decay, yielding \adamw{}.

\subsection{AdamW}
\label{subsec:adamw}

\adamw{} modifies \adam{} by decoupling weight decay from the adaptive gradient update~\citep{loshchilov2019decoupled}. In the coupled formulation, an $\ell_2$ penalty contributes $\lambda \theta_t$ to the gradient before adaptive scaling. This means that the regularization term is itself divided by the adaptive denominator $\sqrt{\hat{v}_t}+\epsilon$, producing coordinate-dependent regularization. \adamw{} instead applies weight decay directly to the parameters:
\begin{equation}
    \theta_{t+1}
    = (1 - \eta_t \lambda)\theta_t
    - \eta_t \frac{\hat{m}_t}{\sqrt{\hat{v}_t}+\epsilon}.
\end{equation}
This decoupling allows the learning rate and weight decay to be tuned more independently.

\adamw{} is the standard reference optimizer for Transformer-based LLMs. Its practical success comes from several properties - 1.\ coordinate-wise adaptivity handles heterogeneous gradient scales, 2.\ momentum smooths noisy minibatch gradients, 3.\ decoupled weight decay provides a stable and interpretable regularization mechanism, 4.\ \adamw{} is widely implemented, optimized, and integrated into distributed-training frameworks.

At the same time, \adamw{} is memory intensive because it stores full first- and second-moment
states, as detailed in \Cref{subsec:bg_optimizer_memory}. This cost is a central reason that
many recent optimizers attempt to reduce, compress, factorize, or restructure moment
estimates while retaining \adamw{}-like behavior.

\subsection{Practical AdamW Design Choices in LLM Training}
\label{subsec:practical_adamw}

The mathematical update rule for \adamw{} does not fully specify a practical LLM training recipe. Several implementation and tuning choices strongly affect performance.

\paragraph{Learning-rate schedule.}
LLM training commonly uses a warmup period followed by cosine decay, linear decay, or inverse-square-root decay. Warmup prevents unstable early updates when moment estimates are poorly calibrated. The optimal schedule interacts with batch size, model scale, dataset size, and training duration.

\paragraph{Momentum coefficients.}
Typical \adamw{} settings use $\beta_1$ near $0.9$ and $\beta_2$ between $0.95$ and $0.999$, although the best values vary by model and training regime. Smaller $\beta_2$ values make the second-moment estimate more responsive to recent gradients, while larger values provide smoother estimates. The choice of $\beta_2$ can affect both stability and generalization.

\paragraph{Weight decay and parameter exclusions.}
In Transformer training, weight decay is often not applied uniformly to all parameters. Biases, normalization parameters, and sometimes embeddings may be excluded from weight decay. These choices matter because such parameters have different scale and regularization behavior than large dense matrices.

\paragraph{Gradient clipping.}
Gradient clipping limits the norm of updates and is often used to prevent rare instability. This is particularly important during early training, when gradients can be large, and in mixed-precision settings, where numerical overflow may occur.

\paragraph{Precision and implementation.}
The moment estimates in \adamw{} are often maintained in fp32 even when model weights and activations use fp16 or bf16. Fused optimizer kernels, optimizer-state sharding, and low-precision variants can significantly affect wall-clock efficiency and memory use. As a result, two implementations of the same mathematical optimizer may have different practical performance.

These details imply that optimizer comparisons should not treat \adamw{} as a single fixed baseline. A weakly tuned \adamw{} baseline can exaggerate the apparent gains of a new optimizer. Conversely, a carefully tuned \adamw{} baseline may reduce or eliminate reported advantages. This issue has been emphasized in recent empirical studies of LLM optimizers~\citep{zhao2025deconstructing,semenov2025benchmarking}.

\subsection{Adafactor}
\label{subsec:adafactor}

\adafactor{} was designed to provide adaptive learning rates with sublinear memory cost~\citep{shazeer2018adafactor}. Its key observation is that many neural-network parameters are matrices. For a matrix parameter $W \in \R^{m \times n}$, a full second-moment accumulator would require $mn$ entries. \adafactor{} instead approximates the second-moment matrix using row and column statistics. If $V_t \in \R^{m \times n}$ denotes the full second-moment estimate, \adafactor{} stores vectors
\begin{align}
    r_t &\in \R^m, \\
    c_t &\in \R^n,
\end{align}
corresponding to row-wise and column-wise averages. These are combined to form an approximate factored preconditioner.

At a high level, the factored approximation reduces memory from $O(mn)$ to $O(m+n)$ for matrix-shaped parameters. This is particularly attractive for Transformer layers, which contain large projection matrices. \adafactor{} can therefore be understood as a memory-efficient adaptive optimizer that exploits parameter shape without fully abandoning coordinate-wise scaling.

\adafactor{} is especially relevant in memory-constrained training and in sequence-to-sequence Transformer models. However, it also changes optimization behavior relative to \adamw{}. The factored preconditioner is an approximation to the full diagonal second moment, and additional design choices---such as whether to use momentum, relative step sizes, update clipping, and parameter scaling---affect performance. In LLM settings, \adafactor{} is an important baseline for evaluating whether newer memory-efficient optimizers provide benefits beyond classical factored adaptivity.

\subsection{Comparing AdamW and Adafactor}
\label{subsec:adamw_adafactor_comparison}

\adamw{} and \adafactor{} represent two different points in the adaptive-optimization design space. \adamw{} stores full first- and second-moment estimates, giving each coordinate its own adaptive statistics. \adafactor{} reduces second-moment memory by using factored statistics for matrix parameters. The trade-off is therefore between fidelity and memory.

\begin{table}[t]
\centering
\small
\begin{tabular}{p{0.22\linewidth}p{0.32\linewidth}p{0.32\linewidth}}
\toprule
\textbf{Property} & \textbf{AdamW} & \textbf{Adafactor} \\
\midrule
Primary motivation & Robust adaptive optimization with decoupled weight decay & Adaptive optimization with reduced memory \\
Second-moment state & Full vector of size $d$ & Factored for matrix parameters \\
Memory cost & High: typically two moment vectors & Lower for large matrices \\
Parameter-structure usage & Mostly coordinate-wise & Exploits matrix shape \\
Common role in LLMs & Default baseline for pretraining and fine-tuning & Memory-efficient adaptive alternative \\
Main limitation & Optimizer-state memory & Approximate preconditioning and tuning sensitivity \\
\bottomrule
\end{tabular}
\caption{Qualitative comparison of \adamw{} and \adafactor{} as adaptive optimizers for LLMs.}
\label{tab:adamw_adafactor}
\end{table}

In practice, choosing between \adamw{} and \adafactor{} depends on the training regime. If memory is abundant and the goal is a strong, well-understood baseline, \adamw{} is difficult to displace. If optimizer-state memory is the binding constraint, \adafactor{} can provide substantial savings. The broader lesson is that LLM optimizer design is often a multi-objective problem: the best optimizer is not necessarily the one with the lowest loss per step, but the one that achieves the best trade-off among loss, memory, stability, and throughput.

\subsection{Limitations of Adaptive Diagonal Optimizers}
\label{subsec:limitations_adaptive_diagonal}

Despite their success, adaptive diagonal optimizers have several limitations that motivate the methods surveyed in later sections.

\paragraph{Memory overhead.}
Full diagonal adaptivity requires storing per-parameter statistics. For \adamw{}, first- and second-moment states can dominate memory usage. This motivates memory-efficient optimizers, quantized states, factorized states, and fused update methods.

\paragraph{Coordinate-wise geometry.}
Adaptive diagonal methods treat each coordinate independently. They do not directly model correlations between parameters or exploit matrix structure. Curvature-aware, low-rank, and matrix-based optimizers can be viewed as attempts to move beyond this coordinate-wise geometry.

\paragraph{Hyperparameter sensitivity.}
Adaptive optimizers introduce several important hyperparameters, including learning rate, $\beta_1$, $\beta_2$, $\epsilon$, weight decay, clipping thresholds, and schedule parameters. Reported optimizer improvements can depend strongly on how these hyperparameters are tuned.

\paragraph{Scale dependence.}
An optimizer that performs well at small scale may not retain its advantage at larger model sizes or longer token budgets. This is particularly important for LLMs, where small-scale experiments are often used to justify methods intended for much larger training runs.

\paragraph{Implementation dependence.}
The practical performance of adaptive optimizers depends on fused kernels, sharding strategies, precision choices, and framework support. Mathematical update rules alone do not determine wall-clock efficiency.

\subsection{Summary}
\label{subsec:adaptive_summary}

Classical first-order methods provide low-memory baselines, while adaptive diagonal methods provide the robustness needed for large-scale Transformer training. \adamw{} remains the dominant LLM optimizer because it combines momentum, coordinate-wise adaptivity, and decoupled weight decay in a well-supported and empirically strong package. However, its memory cost and coordinate-wise geometry have motivated a broad range of alternatives. \adafactor{} demonstrates one important direction: retaining adaptive behavior while exploiting matrix shape to reduce memory. The next section turns to large-batch and distributed-training optimizers, where the optimizer is designed around scaling hardware utilization and stabilizing training at large batch sizes.
\section{Large-Batch and Distributed-Training Optimizers}
\label{sec:largebatch}

Large-scale language models are trained on distributed accelerator clusters where throughput depends strongly on batch size, communication efficiency, and memory layout. Increasing the batch size can improve hardware utilization by amortizing communication and increasing parallelism, but it also changes the optimization problem. Large batches reduce gradient noise, alter the relationship between learning rate and stability, and can degrade generalization if the optimizer is not adjusted carefully. This section reviews large-batch and distributed-training optimizers, with emphasis on layer-wise adaptive methods such as \lamb{} and on the systems-level role of optimizers in LLM training.

\subsection{Why Large-Batch Training Matters}
\label{subsec:why_large_batch}

In data-parallel training, each accelerator processes a local minibatch, gradients are aggregated across workers, and the optimizer updates a synchronized copy of the model. If the global batch size is too small, hardware may be underutilized and communication overhead may dominate. Increasing the global batch size can improve throughput, but it changes the stochastic gradient estimate
\begin{equation}
    g_t = \frac{1}{B}\sum_{i=1}^{B} \nabla_\theta \ell(x_i; \theta_t),
\end{equation}
where $B$ is the global batch size. Larger $B$ reduces the variance of $g_t$, but it also reduces the implicit noise that can help optimization and generalization.

A common heuristic is to increase the learning rate as the batch size grows. However, simple linear scaling can become unstable beyond a critical batch size. The optimizer must then compensate for changes in gradient noise, layer-wise scale, and update magnitude. This motivates large-batch optimizers that normalize or adapt updates at the layer level.

For LLMs, batch size is often measured in tokens rather than examples. The effective batch size depends on the number of sequences, sequence length, gradient accumulation steps, and number of data-parallel workers. Long-context training further complicates this relationship because increasing context length changes both memory use and the number of tokens per batch.

\subsection{Gradient Noise Scale and Critical Batch Size}
\label{subsec:gradient_noise_scale}

The benefit of increasing batch size is limited by the gradient noise scale. Below a problem-dependent critical batch size, larger batches can improve throughput with little loss in optimization efficiency. Above that point, additional batch size yields diminishing returns because the optimizer receives a more accurate gradient estimate but does not make proportionally more progress per step.

This perspective is important for LLM optimizer comparisons. A method may appear faster because it supports a larger batch size, but the relevant question is whether it improves loss as a function of wall-clock time, tokens, or compute. If the batch size exceeds the efficient regime, token efficiency may degrade even if hardware throughput improves. Conversely, a memory-efficient optimizer may enable larger batches, but that only helps if the resulting batch size remains useful for optimization.

Large-batch optimization is therefore inseparable from training economics. The optimal batch size depends on model size, dataset size, sequence length, hardware topology, communication bandwidth, and training duration. Optimizer design interacts with all of these variables.

\subsection{Layer-Wise Adaptive Rate Scaling}
\label{subsec:lars}

Layer-wise adaptive rate scaling (LARS) was introduced to stabilize large-batch training by scaling updates according to the ratio between parameter norms and gradient norms. For a layer or parameter group $\theta^{(l)}$, a simplified trust-ratio update has the form
\begin{equation}
    r_t^{(l)} = \frac{\norm{\theta_t^{(l)}}}{\norm{u_t^{(l)}} + \epsilon},
\end{equation}
where $u_t^{(l)}$ is the proposed update direction for layer $l$. The layer update is then scaled by $r_t^{(l)}$:
\begin{equation}
    \theta_{t+1}^{(l)} = \theta_t^{(l)} - \eta_t r_t^{(l)} u_t^{(l)}.
\end{equation}
The intuition is that each layer should receive an update whose norm is proportional to the norm of its parameters. This prevents layers with unusually large gradients from taking disproportionately large steps and helps maintain stable training at large batch sizes.

Although LARS is more commonly associated with convolutional networks and large-batch vision training, its layer-wise normalization principle influenced later optimizers for large-scale Transformer training, especially \lamb{}.

\subsection{LAMB}
\label{subsec:lamb}

\lamb{} combines Adam-style adaptivity with layer-wise trust ratios~\citep{you2020large}. It was introduced to enable very large-batch training of BERT. The optimizer first computes an Adam-like update direction,
\begin{equation}
    u_t^{(l)} = \frac{\hat{m}_t^{(l)}}{\sqrt{\hat{v}_t^{(l)}} + \epsilon} + \lambda \theta_t^{(l)},
\end{equation}
where $\lambda$ denotes weight decay. It then computes a trust ratio,
\begin{equation}
    r_t^{(l)} = \frac{\norm{\theta_t^{(l)}}}{\norm{u_t^{(l)}}},
\end{equation}
with safeguards for zero norms in practical implementations. The update is
\begin{equation}
    \theta_{t+1}^{(l)} = \theta_t^{(l)} - \eta_t r_t^{(l)} u_t^{(l)}.
\end{equation}

The trust ratio makes the effective learning rate layer-dependent. This can stabilize training when very large batches alter gradient statistics. \lamb{} is therefore not merely an adaptive optimizer; it is an optimizer designed around the interaction between update scale and distributed large-batch training.

In the context of LLMs, \lamb{} is important for two reasons. First, it demonstrates that optimizer design can target hardware efficiency by enabling larger stable batches. Second, it highlights a broader design pattern: global scalar learning rates may be insufficient when different layers require different update magnitudes. This idea reappears in parameter-group-specific optimizers and matrix-aware methods.

\subsection{Large-Batch Training in LLM Pretraining}
\label{subsec:large_batch_llm_pretraining}

LLM pretraining differs from earlier large-batch settings in several ways. First, model sizes are much larger, increasing the memory and communication cost of optimizer states. Second, training often uses long token sequences, making activation memory a major constraint. Third, compute budgets are measured in tokens and floating-point operations over long training horizons, so small differences in optimizer efficiency can have large cost implications.

The global batch size in LLM training is usually determined by a combination of sequence length, microbatch size, gradient accumulation, and number of data-parallel workers. Increasing any of these quantities may improve throughput but can affect optimization. For example, increasing gradient accumulation increases the effective batch size without increasing communication frequency, while increasing data-parallel workers increases communication demands. Optimizers that support stable large-batch training can therefore improve system utilization, but only if the optimization dynamics remain favorable.

\subsection{Distributed Optimizer State and Sharding}
\label{subsec:distributed_optimizer_state}

As summarized in \Cref{tab:optimizer_memory_background}, adaptive optimizer states can be
partitioned across devices using ZeRO-style or fully sharded data-parallel methods. For
large-batch optimizers, this matters because the practical cost of the optimizer depends not
only on its update rule, but also on how its states, gradients, and parameters are distributed
across the training system.

This point is especially important when comparing \adamw{} with proposed alternatives. A new optimizer may reduce per-parameter state in a single-device analysis, but the benefit may shrink when \adamw{} states are already sharded. Conversely, a memory-efficient optimizer may allow larger models or longer contexts even under sharding. The relevant metric is not only state size, but the end-to-end system configuration enabled by that state size.

\subsection{Communication Costs}
\label{subsec:communication_costs}

In distributed training, communication can occur at several stages: gradient synchronization, parameter synchronization, optimizer-state sharding, checkpointing, and pipeline or tensor-parallel communication. Optimizers can affect these costs in several ways.

First, optimizers that require additional statistics may need extra synchronization. Second, matrix-structured or curvature-aware optimizers may require communication of preconditioners or sufficient statistics. Third, low-bit or compressed optimizer states may reduce memory but not necessarily reduce gradient communication. Fourth, optimizer-state checkpointing can become expensive for long training runs, particularly for \adamw{}-style methods with large moment vectors.

Large-batch training can reduce the frequency of synchronization per token by increasing local computation per communication round. However, increasing batch size can also reduce token efficiency. The best optimizer for a distributed setting is therefore the one that optimizes the joint objective of convergence, memory, and communication.

\subsection{Interaction with Learning-Rate Schedules}
\label{subsec:largebatch_lr_schedules}

Large-batch training is sensitive to learning-rate schedules. Warmup is often essential because the optimizer initially has poor moment estimates and the model parameters may be in a fragile regime. For very large batches, the learning rate may need to be scaled, but the scaling rule depends on optimizer, model size, and training phase.

Layer-wise optimizers such as \lamb{} introduce an additional scale factor through the trust ratio. This means the effective step size for a layer is
\begin{equation}
    \eta_t^{(l)} = \eta_t r_t^{(l)}.
\end{equation}
As a result, the global schedule and layer-wise trust ratios interact. This interaction complicates optimizer comparisons because two methods with the same nominal learning rate may produce very different effective update magnitudes.

\subsection{Practical Role of Large-Batch Optimizers in Modern LLM Training}
\label{subsec:practical_role_large_batch}

Although \lamb{} was influential for large-batch BERT training, many modern LLM pretraining pipelines still rely on carefully tuned \adamw{} combined with distributed training techniques. This does not mean large-batch optimizers are irrelevant. Rather, their role has shifted. They provide a conceptual and practical framework for understanding how update scale, batch size, and hardware utilization interact.

Large-batch methods are most relevant when the training setup is communication-limited or when increasing the batch size substantially improves throughput. They are less compelling if the model already operates near the critical batch size, if optimizer memory rather than synchronization is the main bottleneck, or if larger batches reduce final quality. Therefore, the usefulness of large-batch optimizers depends on the complete training regime.

\subsection{Evaluation Considerations}
\label{subsec:largebatch_evaluation}

Large-batch optimizers should be evaluated with particular attention to the relationship
between batch size, throughput, and convergence. In addition to the general benchmarking
criteria discussed in \Cref{sec:benchmarking}, papers should report global batch size in
tokens, microbatch size, gradient accumulation steps, learning-rate scaling rules, warmup
schedule, and distributed-training configuration. Without these details, it is difficult to
determine whether an optimizer improves the underlying optimization process or merely
benefits from a more favorable systems configuration.

\subsection{Summary}
\label{subsec:largebatch_summary}

Large-batch and distributed-training optimizers address the interaction between optimization and hardware efficiency. Methods such as \lamb{} use layer-wise trust ratios to stabilize large-batch training, while distributed systems reduce the memory cost of adaptive optimizers through sharding. For LLMs, the central issue is not simply whether a larger batch can be used, but whether the optimizer improves the trade-off among token efficiency, wall-clock time, memory, communication, and final model quality. The next section turns to memory-efficient optimizers, which target one of the most persistent bottlenecks in LLM training: the cost of storing optimizer states.

\section{Memory-Efficient Optimizers}
\label{sec:memory}

Memory-efficient optimizers target one of the central bottlenecks in LLM training: optimizer-state memory. For adaptive optimizers such as \adamw{}, the first- and second-moment estimates can require multiple additional copies of the model parameters. At billion-parameter scale, this overhead directly limits the maximum model size, sequence length, batch size, and feasibility of full-parameter fine-tuning. This section reviews optimizer designs that reduce memory while attempting to preserve the convergence behavior of adaptive methods.

\subsection{Memory as an Optimization Constraint}
\label{subsec:memory_constraint}

The memory footprint of LLM training can be decomposed into several components:
\begin{equation}
    M_{\mathrm{total}}
    = M_{\mathrm{params}} + M_{\mathrm{grads}} + M_{\mathrm{opt}} + M_{\mathrm{acts}} + M_{\mathrm{misc}},
\end{equation}
where $M_{\mathrm{params}}$ stores model parameters, $M_{\mathrm{grads}}$ stores gradients, $M_{\mathrm{opt}}$ stores optimizer states, $M_{\mathrm{acts}}$ stores activations, and $M_{\mathrm{misc}}$ includes buffers, fragmentation, communication workspaces, and framework overhead. Optimizer design primarily targets $M_{\mathrm{opt}}$, but reducing optimizer memory can indirectly allow larger activations, longer contexts, or larger microbatches.

As discussed in \Cref{subsec:bg_optimizer_memory}, \adamw{} stores full first- and
second-moment states, making optimizer memory a major scaling cost. Memory-efficient
optimizers target this term directly by using sublinear, compressed, grouped, fused, or
lower-precision state.

Memory-efficient optimizers are not merely engineering conveniences. They can change the feasible training regime. A memory reduction may allow full-parameter fine-tuning instead of parameter-efficient fine-tuning, a longer context window instead of shorter sequences, or a larger model under the same hardware budget. Thus, memory-efficient optimization should be evaluated at the system level: the relevant question is not only whether the optimizer matches \adamw{} at fixed model size, but also what model, context, and batch configurations it makes possible.

\subsection{Design Patterns for Reducing Optimizer Memory}
\label{subsec:memory_design_patterns}

Memory-efficient optimizers use several recurring design patterns.

\paragraph{State elimination.}
The most direct approach is to remove some optimizer states. Plain \sgd{} stores no moment vectors, and momentum stores only one. However, eliminating adaptivity can harm convergence for LLMs. Many memory-efficient optimizers therefore seek intermediate points between \sgd{} and \adamw{}.

\paragraph{State factorization.}
For matrix-shaped parameters, a full per-coordinate statistic can be approximated by lower-dimensional row and column statistics. \adafactor{} is the canonical example~\citep{shazeer2018adafactor}. Factorization reduces memory from $O(mn)$ to $O(m+n)$ for an $m \times n$ matrix.

\paragraph{State quantization.}
Moment estimates can be stored in lower precision. 8-bit optimizers use block-wise quantization to reduce optimizer-state memory while preserving enough numerical fidelity for training~\citep{dettmers2022optimizers}.

\paragraph{State grouping.}
Instead of assigning a distinct adaptive statistic to every parameter, parameters can be grouped so that multiple parameters share learning-rate statistics. Adam-mini follows this direction by reducing the number of distinct learning rates used by Adam-style optimization~\citep{zhang2025adammini}.

\paragraph{Fused update and gradient handling.}
Some methods reduce memory by avoiding the materialization of intermediate tensors or by fusing gradient computation with parameter updates. \lomo{} is an example designed for memory-efficient full-parameter fine-tuning~\citep{lv2024lomo}.

\paragraph{Projection and low-rank structure.}
Gradient projection methods reduce memory by representing updates in a lower-dimensional subspace. Methods such as \galore{} are discussed in detail in \Cref{sec:lowrank}, but they are also part of the broader memory-efficiency landscape~\citep{zhao2024galore}.

\subsection{Adafactor: Factored Second-Moment Estimation}
\label{subsec:memory_adafactor}

\adafactor{} reduces the memory cost of adaptive learning rates through factored second-moment estimation~\citep{shazeer2018adafactor}. For a matrix parameter $W \in \R^{m \times n}$ with gradient $G_t$, a full second-moment accumulator would store an $m \times n$ matrix. \adafactor{} instead stores row and column accumulators. A simplified version computes
\begin{align}
    R_t &= \beta_2 R_{t-1} + (1-\beta_2) \frac{1}{n}\sum_{j=1}^{n} G_{t,:,j}^2, \\
    C_t &= \beta_2 C_{t-1} + (1-\beta_2) \frac{1}{m}\sum_{i=1}^{m} G_{t,i,:}^2,
\end{align}
where $R_t \in \R^m$ and $C_t \in \R^n$. These statistics are combined to approximate the full second-moment matrix.

The memory benefit is substantial for large matrices. Instead of $mn$ second-moment values, \adafactor{} stores $m+n$ values. Transformer models contain many large projection matrices, making this factorization attractive. However, the approximation is less expressive than a full diagonal accumulator. It assumes that the second-moment structure can be approximated from row and column marginals.

\adafactor{} also includes practical design choices such as update clipping, relative step-size schedules, and optional momentum. These details matter. In some settings, \adafactor{} can work well as a drop-in memory-saving replacement, while in others it requires careful tuning. For LLM optimizer surveys, \adafactor{} is important because it introduced the idea that optimizer state should exploit tensor shape rather than treating every parameter independently.

\subsection{Low-Bit Optimizer States}
\label{subsec:lowbit_optimizer_states}

Quantizing optimizer states reduces memory by storing moments in fewer bits. 8-bit optimizers use block-wise quantization to represent optimizer states compactly while maintaining training stability~\citep{dettmers2022optimizers}. At a high level, a tensor is divided into blocks, each block is quantized using local scaling statistics, and optimizer updates are computed with dequantized or approximately dequantized values.

The appeal of low-bit optimizer states is that they preserve the structure of the original optimizer. Unlike factorization or grouping, quantization does not necessarily change which statistics are tracked; it changes how they are stored. This can make low-bit optimizers easier to interpret as memory-reduced versions of \adamw{}.

The trade-offs are numerical and systems related. Quantization introduces approximation error, and the error distribution can interact with small gradient magnitudes, rare large updates, and mixed-precision arithmetic. Low-bit optimizers may also require specialized kernels to realize wall-clock gains. If quantization reduces memory but adds dequantization overhead, the net benefit depends on the hardware and implementation.

\subsection{Adam-mini: Reducing Redundant Adaptivity}
\label{subsec:adammini}

Adam-mini is motivated by the observation that \adam{} and \adamw{} assign an independent adaptive learning-rate statistic to every parameter, even though some of this adaptivity may be redundant in large neural networks~\citep{zhang2025adammini}. Instead of storing a full second-moment estimate for every coordinate, Adam-mini reduces the number of distinct learning rates by grouping parameters according to structure.

This approach differs from quantization and factorization. Quantization stores the same kind of state in fewer bits. Factorization approximates a full matrix of statistics with row and column summaries. Adam-mini changes the granularity of adaptivity: multiple parameters may share an adaptive statistic. The goal is to preserve the useful part of Adam-style scaling while removing unnecessary per-parameter state.

The conceptual contribution of Adam-mini is that memory-efficient optimization need not simply approximate \adamw{} as closely as possible. Instead, it can ask which parts of \adamw{} are actually necessary for LLM training. If many parameters do not require independent learning-rate estimates, then grouped adaptivity can reduce memory without severely harming convergence.

\subsection{LOMO: Memory-Efficient Full-Parameter Fine-Tuning}
\label{subsec:lomo}

\lomo{} targets full-parameter fine-tuning of large language models under limited resources~\citep{lv2024lomo}. The key motivation is that standard fine-tuning with \adamw{} can be prohibitively memory intensive because it requires gradients and optimizer states for all parameters. Parameter-efficient fine-tuning methods avoid this by updating only a small number of parameters, but they do not perform full-parameter adaptation.

\lomo{} reduces memory by fusing gradient computation and parameter updates, thereby avoiding the need to store all gradients and optimizer states in the conventional way. Conceptually, this shifts optimization from a state-heavy adaptive update to a memory-frugal update procedure suitable for full-model fine-tuning.

The importance of \lomo{} for this survey is its training-regime focus. It is not primarily designed to outperform \adamw{} in large-scale pretraining. Instead, it addresses a different question: how can one update all parameters of a large model when hardware memory is limited? This distinction illustrates why optimizer evaluation must specify the target regime. An optimizer can be valuable because it enables a training mode that would otherwise be infeasible.

\subsection{APOLLO-like SGD-Memory Adaptive Methods}
\label{subsec:apollo_like}

Recent methods such as APOLLO aim to approach \adamw{}-level performance with memory closer to \sgd{}~\citep{han2024apollo}. These approaches are part of a broader trend toward adaptive behavior without full adaptive state. While the precise mechanisms differ across methods, the shared goal is to reduce or eliminate large per-parameter moment tensors while preserving the optimization benefits of adaptive scaling.

This direction is important because it challenges the assumption that two full moment vectors are necessary for strong LLM training. If adaptive behavior can be approximated through lower-memory statistics, projection, grouping, or structured updates, then optimizer memory may no longer scale as a fixed multiple of model size. However, such claims require careful validation across model scales, datasets, and tuning budgets.

\subsection{Memory-Efficient Optimizers and Distributed Sharding}
\label{subsec:memory_and_sharding}

Optimizer-state sharding and memory-efficient optimizers are complementary rather than
substitutes. Sharding reduces per-device memory, while memory-efficient optimizers reduce
the amount of state that must be stored, communicated, and checkpointed in the first place.
Thus, even under ZeRO-style or FSDP-style sharding, reducing optimizer state can enable
larger models, longer contexts, larger microbatches, or fewer required devices.

\subsection{Trade-Offs in Memory-Efficient Optimization}
\label{subsec:memory_tradeoffs}

Memory-efficient optimizers involve several trade-offs.

\paragraph{Memory versus update fidelity.}
Reducing state may make the optimizer less faithful to full \adamw{} adaptivity. The question is whether the lost information matters for the target model and training regime.

\paragraph{Memory versus computation.}
Some methods reduce memory but add computation. For example, quantization may require extra encode/decode operations, and projection methods may require subspace updates.

\paragraph{Memory versus implementation complexity.}
A simple optimizer with high memory may be easier to deploy than a memory-efficient optimizer requiring custom kernels, parameter grouping, or nonstandard distributed support.

\paragraph{Memory versus convergence.}
An optimizer that saves memory but requires more tokens to reach the same loss may or may not be preferable. If the memory savings allow a larger model or longer context, the trade-off can still be favorable.

\paragraph{Pretraining versus fine-tuning.}
The best memory-efficient optimizer may differ by regime. Pretraining emphasizes long-horizon stability and throughput, while fine-tuning may emphasize feasibility on limited hardware and robustness to small datasets.

\subsection{Evaluation Criteria}
\label{subsec:memory_evaluation}

Memory-efficient optimizers require both fixed-model and fixed-resource evaluation. In a
fixed-model setting, the question is whether the method matches \adamw{} on the same
architecture and token budget while using less memory. In a fixed-resource setting, the
question is whether the saved memory enables a larger model, longer context, or larger
batch under the same hardware budget. In addition to the general protocol in
\Cref{sec:benchmarking}, evaluations should report peak memory, optimizer-state size,
checkpoint size, feasible context length, feasible microbatch size, and wall-clock time to a
target loss.

\subsection{Summary}
\label{subsec:memory_summary}

Memory-efficient optimizers reduce one of the most important costs of LLM training: optimizer-state memory. \adafactor{} reduces memory through factorized second-moment estimates; 8-bit optimizers compress states through quantization; Adam-mini reduces redundant adaptivity through grouping; \lomo{} enables full-parameter fine-tuning with reduced memory; and APOLLO-like methods seek \sgd{}-like memory with \adamw{}-level performance. These methods show that optimizer design for LLMs is not only about convergence, but also about what training regimes become feasible. The next section turns to sign-based and discovered optimizers, which modify the update rule itself rather than primarily targeting memory footprint.

\section{Sign-Based and Discovered Optimizers}
\label{sec:sign}

Sign-based and discovered optimizers modify the update rule itself rather than primarily reducing optimizer-state memory or adding curvature information. The central idea is that the precise magnitude of a gradient coordinate may be less important than its direction or sign, especially when combined with momentum, adaptive schedules, and normalization. This section reviews sign-based optimization, the symbolic discovery of \lion{}, and the relevance and limitations of such methods for LLM training.

\subsection{Motivation for Sign-Based Updates}
\label{subsec:sign_motivation}

Standard first-order optimizers update parameters using gradient magnitudes. For example, \sgd{} applies the update $-\eta_t g_t$, and \adamw{} applies a momentum direction rescaled by a second-moment estimate. Sign-based optimizers instead use the sign of a gradient-derived quantity:
\begin{equation}
    \theta_{t+1} = \theta_t - \eta_t \operatorname{sign}(u_t),
\end{equation}
where $u_t$ may be the current gradient, a momentum estimate, or a learned combination of past gradients.

This update has several appealing properties. First, it is invariant to the magnitude of each coordinate, depending only on direction. Second, it can be viewed as a form of aggressive normalization, where every active coordinate receives an update of equal magnitude. Third, sign updates can be communication efficient in distributed settings because signs are easier to compress than full-precision gradients. Fourth, sign-based rules can reduce sensitivity to outlier gradient magnitudes, although they may introduce sensitivity to learning-rate tuning.

For LLM training, sign-based methods are interesting because they challenge the assumption that high-quality optimization requires full coordinate-wise second-moment adaptivity. If signs of momentum-like quantities are sufficient for competitive training, then one may obtain simpler update rules with reduced state or improved efficiency. However, this promise must be evaluated carefully at scale.

\subsection{signSGD}
\label{subsec:signsgd}

A canonical sign-based method is signSGD~\citep{bernstein2018signsgd}. In its simplest form, signSGD updates parameters according to
\begin{equation}
    \theta_{t+1} = \theta_t - \eta_t \operatorname{sign}(g_t).
\end{equation}
In distributed settings, signSGD can be combined with majority vote aggregation, where workers send gradient signs and the global update follows the coordinate-wise majority sign.

The main advantage of signSGD is communication efficiency. Sending one bit per coordinate can substantially reduce communication compared with transmitting full-precision gradients. This is appealing in distributed optimization settings where communication is a bottleneck. However, signSGD also discards magnitude information, which can be important for ill-conditioned problems or for parameters with very different sensitivities.

For LLMs, plain signSGD is mainly a conceptual baseline rather than a standard training method. Modern sign-based optimizers typically combine sign updates with momentum, weight decay, and carefully tuned learning-rate schedules. The lesson from signSGD is that sign information alone can sometimes produce useful descent directions, but practical LLM optimization requires additional mechanisms for stability and scale.

\subsection{Symbolic Discovery of Optimizers}
\label{subsec:symbolic_discovery}

Optimizer design has traditionally relied on mathematical analysis, intuition, and empirical tuning. Symbolic optimizer discovery offers a different approach: search over a space of possible update rules and select algorithms that perform well on training tasks. \citet{chen2023symbolic} use symbolic discovery to identify new optimization algorithms, including \lion{}, a simple sign-based optimizer.

The discovered optimizer is notable because it is compact and interpretable. Rather than producing a large learned optimizer network, the search yields a human-readable update rule. This makes the resulting method easier to analyze, implement, and compare with hand-designed optimizers.

Symbolic discovery raises an important methodological question for LLM optimization. If an optimizer is discovered on proxy tasks or smaller models, will it transfer to full-scale LLM training? The answer depends on whether the search tasks capture the optimization geometry, gradient statistics, scale, and systems constraints of the target setting. For this reason, discovered optimizers require especially careful validation across model sizes and training regimes.

\subsection{Lion}
\label{subsec:lion}

\lion{} is a sign-based optimizer discovered through symbolic search~\citep{chen2023symbolic}. It maintains a momentum-like exponential moving average of gradients, but uses the sign of a linear combination of the current gradient and momentum for the parameter update. A simplified form is
\begin{align}
    u_t &= \beta_1 m_{t-1} + (1 - \beta_1) g_t, \\
    \theta_{t+1} &= \theta_t - \eta_t \left( \operatorname{sign}(u_t) + \lambda \theta_t \right), \\
    m_t &= \beta_2 m_{t-1} + (1 - \beta_2) g_t,
\end{align}
where $m_t$ is the momentum state and $\lambda$ is a weight-decay coefficient. Different presentations may vary in notation, but the defining feature is that the update direction uses a sign operation applied to a momentum-like quantity.

Compared with \adamw{}, \lion{} does not maintain a second-moment vector. This can reduce optimizer-state memory from two moment vectors to one. It also changes the update geometry: \adamw{} scales coordinates by an estimate of gradient variance, whereas \lion{} normalizes update magnitudes through the sign operation.

\lion{} is attractive for LLMs for three reasons. First, it is simple. Second, it can reduce optimizer-state memory relative to \adamw{}. Third, it has shown strong empirical performance across several deep-learning settings. However, its performance can depend strongly on learning-rate and weight-decay tuning. Since sign updates have fixed coordinate-wise magnitudes, the global learning rate plays a particularly important role in determining update scale.

\subsection{Relationship Between Lion and AdamW}
\label{subsec:lion_adamw_relationship}

\lion{} and \adamw{} can be contrasted through their treatment of gradient magnitude. \adamw{} uses the ratio
\begin{equation}
    \frac{\hat{m}_t}{\sqrt{\hat{v}_t}+\epsilon},
\end{equation}
where the denominator provides coordinate-wise adaptivity. \lion{} instead uses
\begin{equation}
    \operatorname{sign}(u_t),
\end{equation}
where $u_t$ is a momentum-like direction. Thus, \adamw{} rescales each coordinate continuously, while \lion{} maps each coordinate to a discrete direction in $\{-1,0,1\}$.

This distinction has several consequences. \adamw{} can take smaller steps in coordinates with large historical gradient variance, while \lion{} cannot distinguish such coordinates except through the sign of the update direction. On the other hand, \lion{} avoids storing the second-moment vector and may be less affected by noisy magnitude estimates. In effect, \lion{} trades coordinate-wise adaptive scaling for a simpler sign-normalized update.

The practical comparison is therefore empirical and regime dependent. If gradient magnitudes contain essential information, \adamw{} may be preferable. If signs of momentum provide sufficient information and memory or simplicity matters, \lion{} may be attractive. This trade-off should be evaluated under matched tuning budgets and realistic LLM training settings.

\subsection{Learning-Rate and Weight-Decay Sensitivity}
\label{subsec:sign_lr_sensitivity}

Sign-based optimizers can be sensitive to learning-rate choices. Because each coordinate update has approximately fixed magnitude, the learning rate directly controls the coordinate-wise step size. In \adamw{}, the adaptive denominator can dampen large-gradient coordinates; in sign-based methods, such damping is absent unless introduced by other mechanisms.

Weight decay also interacts differently with sign updates. Decoupled weight decay remains important, but the relative scale of decay and sign updates can differ from \adamw{}. As a result, hyperparameters that work for \adamw{} may not transfer directly to \lion{}. A fair comparison must therefore tune learning rate, weight decay, momentum coefficients, schedule, and clipping separately for each optimizer.

This issue is central to evaluating sign-based optimizers for LLMs. A new optimizer may appear better than \adamw{} if \adamw{} is under-tuned or if the new optimizer receives a more favorable schedule. Conversely, an optimizer may appear worse if it is evaluated with hyperparameters inherited from \adamw{} without adjustment.

\subsection{Memory and Communication Considerations}
\label{subsec:sign_memory_communication}

Sign-based optimizers can reduce optimizer-state memory when they avoid second-moment accumulators. For example, a momentum-based sign optimizer may store only one auxiliary vector rather than the two vectors used by \adamw{}. This places sign-based methods between momentum \sgd{} and \adamw{} in memory cost.

In distributed settings, sign information can also be compressed more easily than full-precision gradients. However, this potential advantage does not automatically translate into faster LLM training. Modern LLM systems are often bottlenecked by a combination of computation, activation memory, optimizer-state memory, and communication. If gradient synchronization remains full precision or if sign compression is not supported by the training stack, the theoretical communication advantage may not be realized.

Thus, sign-based optimizers should be evaluated both algorithmically and systemically. Their memory savings are clearest when compared with full-state \adamw{}. Their communication benefits depend on whether the implementation actually communicates compressed signs or merely uses signs inside the local update rule.

\subsection{Empirical Evaluation Challenges}
\label{subsec:sign_empirical_challenges}

Sign-based and discovered optimizers require especially careful tuning because their effective
update scale differs from \adamw{}. Learning rate, weight decay, momentum coefficients,
and schedules should therefore be tuned separately rather than inherited from \adamw{}.
Moreover, optimizers discovered on proxy tasks should be validated across model scales and
training horizons before being treated as LLM-scale replacements. We discuss the general
benchmarking protocol for optimizer comparisons in \Cref{sec:benchmarking}.

\subsection{Role in the Broader Optimizer Landscape}
\label{subsec:sign_role_landscape}

Sign-based and discovered optimizers occupy an important position in the LLM optimizer landscape. They show that competitive optimization may not require full second-moment adaptivity in every setting. This observation supports a broader trend: revisiting which components of \adamw{} are essential and which can be simplified, compressed, or replaced.

At the same time, sign-based methods do not fully address all LLM optimizer challenges. They may reduce memory, but not as aggressively as factorized or low-rank methods. They may simplify updates, but do not explicitly exploit matrix structure. They may improve some empirical results, but require careful tuning and validation. Their main contribution is therefore conceptual as much as practical: they expand the design space beyond coordinate-wise adaptive scaling.

\subsection{Summary}
\label{subsec:sign_summary}

Sign-based and discovered optimizers replace or simplify magnitude-sensitive updates using sign-normalized directions. signSGD demonstrates the basic principle, while \lion{} shows how symbolic search can produce a compact momentum-based sign optimizer. These methods can reduce optimizer-state memory and challenge the necessity of full second-moment adaptivity. However, their value for LLM training depends on fair hyperparameter tuning, scale-aware evaluation, and high-quality implementation. The next section considers a different route beyond \adamw{}: curvature-aware and second-order optimization, which seeks better update geometry by approximating curvature rather than discarding magnitude information.

\section{Curvature-Aware and Second-Order Optimizers}
\label{sec:curvature}

Adaptive first-order optimizers such as \adamw{} use diagonal statistics of past gradients to rescale coordinates. This diagonal scaling is computationally efficient and robust, but it does not explicitly model interactions among parameters. Curvature-aware optimizers attempt to improve update geometry by incorporating information about the local curvature of the loss. In principle, second-order information can improve conditioning and reduce the number of optimization steps required to reach a target loss. In practice, exact second-order optimization is infeasible for LLMs, so modern methods use diagonal, block-wise, Kronecker-factored, or matrix-structured approximations.

This section reviews curvature-aware optimization for LLMs. We first motivate second-order preconditioning, then discuss why exact Newton-style methods are impractical at LLM scale. We then review Shampoo-style matrix preconditioning, \sophia{}, and SOAP-like structured methods. Finally, we discuss the trade-off between token efficiency and per-step overhead.

\subsection{From Diagonal Adaptivity to Curvature}
\label{subsec:curvature_motivation}

Consider minimizing a twice-differentiable loss $\mathcal{L}(\theta)$. A local quadratic approximation around $\theta_t$ is
\begin{equation}
    \mathcal{L}(\theta_t + \Delta)
    \approx
    \mathcal{L}(\theta_t)
    + g_t^\top \Delta
    + \frac{1}{2}\Delta^\top H_t \Delta,
\end{equation}
where $g_t = \nabla_\theta \mathcal{L}(\theta_t)$ and $H_t = \nabla_\theta^2 \mathcal{L}(\theta_t)$ is the Hessian. If $H_t$ is positive definite, the Newton update is
\begin{equation}
    \Delta_t = - H_t^{-1} g_t.
\end{equation}
This update rescales and rotates the gradient according to the local curvature of the loss.

In high-dimensional neural networks, exact Newton updates are impractical. The Hessian has $d^2$ entries for $d$ parameters, and computing, storing, communicating, and inverting it is infeasible for modern LLMs. Nevertheless, the motivation remains important: if the loss is ill-conditioned, a curvature-aware preconditioner may make optimization more efficient than a purely first-order update.

Diagonal adaptive methods can be interpreted as crude curvature or scale approximations. \adamw{} uses the square root of an exponential moving average of squared gradients as a diagonal preconditioner. This captures per-coordinate gradient scale, but not cross-coordinate correlations. Curvature-aware methods seek richer approximations while remaining computationally feasible.

\subsection{General Preconditioned Updates}
\label{subsec:preconditioned_updates}

Many optimizers can be written as preconditioned gradient methods:
\begin{equation}
    \theta_{t+1} = \theta_t - \eta_t P_t g_t,
\end{equation}
where $P_t$ is a preconditioner. For \sgd{}, $P_t = I$. For diagonal adaptive methods, $P_t$ is diagonal. For second-order methods, $P_t$ approximates $H_t^{-1}$ or a related curvature inverse.

The design of $P_t$ determines the trade-off among conditioning, memory, and computation. A richer preconditioner can better account for parameter interactions, but it may be expensive to store and apply. Curvature-aware LLM optimizers therefore rely on structure: diagonal Hessian estimates, block-wise approximations, Kronecker factors, tensor dimensions, or matrix-specific preconditioners.

\subsection{Shampoo and Tensor Preconditioning}
\label{subsec:shampoo}

Shampoo is a preconditioned stochastic optimization method that exploits tensor structure~\citep{gupta2018shampoo}. For a matrix-shaped parameter $W \in \R^{m \times n}$ with gradient $G_t$, Shampoo maintains preconditioner statistics along the matrix dimensions. A simplified version accumulates
\begin{align}
    L_t &= L_{t-1} + G_t G_t^\top, \\
    R_t &= R_{t-1} + G_t^\top G_t,
\end{align}
where $L_t \in \R^{m \times m}$ and $R_t \in \R^{n \times n}$. The gradient is then preconditioned using inverse matrix roots:
\begin{equation}
    \widetilde{G}_t = L_t^{-1/4} G_t R_t^{-1/4}.
\end{equation}
The parameter update uses $\widetilde{G}_t$ rather than $G_t$.

The key idea is to approximate curvature along tensor dimensions rather than storing a full $mn \times mn$ preconditioner. This makes the method more scalable than full second-order optimization while capturing more structure than diagonal adaptivity. Extensions and scalable implementations of Shampoo have explored practical second-order optimization for deep learning~\citep{anil2021scalable}.

For LLMs, Shampoo-style methods are conceptually attractive because Transformer parameters are dominated by large matrices. However, applying matrix inverse roots can be expensive, especially for large dimensions. Practical implementations often update preconditioners less frequently, use numerical approximations, block large matrices, or restrict the method to selected parameter groups. The main question is whether improved conditioning compensates for this overhead at LLM scale.

\subsection{Sophia}
\label{subsec:sophia}

\sophia{} is a scalable stochastic second-order optimizer designed for language-model pretraining~\citep{liu2024sophia}. Its central idea is to use a lightweight diagonal estimate of curvature to precondition the update. Instead of storing or inverting a full Hessian, \sophia{} maintains an estimate $h_t$ of diagonal curvature and uses it to scale a momentum-like direction.

A simplified form of the update is
\begin{align}
    m_t &= \beta_1 m_{t-1} + (1-\beta_1) g_t, \\
    h_t &\approx \operatorname{diag}(H_t), \\
    \theta_{t+1} &= \theta_t - \eta_t \cdot \operatorname{clip}\left(\frac{m_t}{h_t + \epsilon}, \rho\right),
\end{align}
where $\rho$ denotes a clipping threshold. The exact estimator and update details vary by variant, but the defining feature is the use of a diagonal Hessian-like quantity rather than only squared-gradient moments.

The motivation differs from \adamw{}. \adamw{} estimates second moments of stochastic gradients; \sophia{} aims to estimate curvature more directly. This distinction matters because gradient variance and curvature are related but not identical. A curvature estimate may provide a better preconditioner when the objective is ill-conditioned.

\sophia{} is especially important for this survey because it is explicitly designed for language-model pretraining. It represents a practical attempt to bring second-order information into LLM optimization without incurring the cost of full Newton or full-matrix preconditioning. Its reported benefits should nevertheless be interpreted through the same evaluation lens as other optimizers: scale, tuning budget, wall-clock cost, memory overhead, and implementation details.

\subsection{SOAP-like Structured Preconditioning}
\label{subsec:soap_like}

SOAP-like methods combine adaptive optimization with structured preconditioning. At a high level, they attempt to retain the practical advantages of Adam-style updates while incorporating richer matrix or tensor geometry. Such methods often use approximate second-order statistics, orthogonal transformations, or preconditioning in a rotated basis.

The attraction of SOAP-like methods for LLMs is that they may improve conditioning for large dense matrices without requiring a full Hessian. They sit between diagonal methods such as \adamw{} and more expensive matrix-preconditioning methods such as full Shampoo. Their practical effectiveness depends on whether the additional structure can be computed and applied efficiently on accelerator hardware.

Because the literature on structured preconditioning for LLMs is developing rapidly, these methods should be discussed with particular care. Important questions include whether gains persist under strong \adamw{} baselines, whether the methods scale to multi-billion-parameter models, and whether the preconditioning overhead is offset by fewer required training tokens or faster wall-clock convergence.

\subsection{Curvature Estimation Choices}
\label{subsec:curvature_estimation_choices}

Curvature-aware methods differ in what quantity they estimate and how they estimate it.

\paragraph{Squared-gradient statistics.}
Adaptive optimizers such as \adamw{} use exponential moving averages of squared gradients. These are not Hessian estimates, but they provide scale information that can function as a diagonal preconditioner.

\paragraph{Diagonal Hessian estimates.}
Methods such as \sophia{} approximate the diagonal of the Hessian or related curvature quantities. This can provide a more direct estimate of local curvature while maintaining linear memory cost.

\paragraph{Matrix or tensor statistics.}
Shampoo-style methods accumulate statistics along tensor dimensions, such as $GG^\top$ and $G^\top G$. These capture correlations within matrix rows and columns.

\paragraph{Block-wise approximations.}
Large matrices can be split into blocks to reduce the cost of preconditioning. This trades off fidelity and scalability.

\paragraph{Update frequency.}
Curvature statistics may be updated less frequently than gradients. Less frequent updates reduce cost but may make the preconditioner stale.

Each design choice affects memory, computation, numerical stability, and implementation complexity.

\subsection{Memory and Compute Costs}
\label{subsec:curvature_costs}

Curvature-aware optimizers often require more memory or computation than first-order methods. A diagonal curvature estimate has memory cost comparable to an additional vector. Matrix preconditioners can be substantially more expensive. For a matrix parameter $W \in \R^{m \times n}$, Shampoo-style statistics require $O(m^2+n^2)$ storage rather than $O(mn)$ storage. This may be efficient for some shapes but expensive for others.

The compute cost can also be significant. Matrix inverse roots, eigendecompositions, Newton--Schulz iterations, or preconditioner updates may add overhead. In distributed training, curvature statistics may require synchronization or careful partitioning. These costs must be compared with the potential reduction in the number of training steps or tokens.

A curvature-aware optimizer is beneficial when
\begin{equation}
    \text{extra cost per step} < \text{cost saved by faster convergence},
\end{equation}
under the relevant training budget. This inequality depends on hardware, implementation, model scale, and target loss.

\subsection{Curvature-Aware Optimization and LLM Scale}
\label{subsec:curvature_llm_scale}

Scaling curvature-aware methods to LLMs introduces several challenges.

\paragraph{Parameter size.}
LLMs contain very large matrices, making full matrix preconditioning expensive.

\paragraph{Distributed partitioning.}
Parameters and optimizer states may be sharded across devices. Structured preconditioners must be compatible with this partitioning.

\paragraph{Mixed precision.}
Curvature estimates can be numerically sensitive. Low-precision training may require careful accumulation and stabilization.

\paragraph{Update frequency.}
Updating curvature estimates every step may be too expensive. Infrequent updates reduce overhead but may reduce preconditioner quality.

\paragraph{Hyperparameter complexity.}
Curvature-aware methods introduce additional hyperparameters, such as damping, clipping, preconditioner decay, update frequency, and block size.

These challenges do not make curvature-aware optimization impractical, but they raise the bar for evaluation. A method must show that improved optimization geometry outweighs additional complexity.

\subsection{Comparison with AdamW}
\label{subsec:curvature_vs_adamw}

\adamw{} can be viewed as a strong diagonal-preconditioning baseline. Curvature-aware methods attempt to improve on it by replacing or augmenting squared-gradient scaling with curvature information. The comparison involves several dimensions:

\begin{table}[t]
\centering
\small
\begin{tabular}{p{0.22\linewidth}p{0.30\linewidth}p{0.34\linewidth}}
\toprule
\textbf{Property} & \textbf{AdamW} & \textbf{Curvature-aware methods} \\
\midrule
Preconditioner & Diagonal squared-gradient statistics & Diagonal Hessian, block, matrix, or tensor curvature approximations \\
Memory cost & Two moment vectors & Ranges from vector-level to matrix-level states \\
Compute overhead & Low to moderate & Moderate to high, depending on approximation \\
Main advantage & Robust, widely supported baseline & Potentially better conditioning and token efficiency \\
Main risk & High memory and coordinate-wise limitation & Overhead, complexity, and tuning sensitivity \\
\bottomrule
\end{tabular}
\caption{Qualitative comparison between \adamw{} and curvature-aware optimizer families.}
\label{tab:curvature_adamw_comparison}
\end{table}

A curvature-aware optimizer should therefore be compared to a well-tuned \adamw{} baseline using both token-based and wall-clock metrics. It is insufficient to report fewer steps to a target loss if each step is much more expensive. Conversely, a method with higher per-step cost may still be valuable if it enables lower total compute or better final quality.

\subsection{Evaluation Guidelines}
\label{subsec:curvature_evaluation}

Evaluations of curvature-aware optimizers should report:
\begin{enumerate}[leftmargin=*]
    \item validation loss as a function of tokens;
    \item validation loss as a function of wall-clock time;
    \item per-step compute overhead;
    \item peak memory usage, including preconditioner states;
    \item preconditioner update frequency;
    \item precision used for curvature statistics;
    \item sensitivity to damping, clipping, and block size;
    \item compatibility with distributed sharding;
    \item downstream performance at fixed compute budget.
\end{enumerate}

These details are necessary because curvature-aware methods often trade cheaper optimization trajectories for more expensive steps. Without full accounting, it is difficult to determine whether they improve practical LLM training.

\subsection{Summary}
\label{subsec:curvature_summary}

Curvature-aware optimizers seek better update geometry than diagonal adaptive methods by approximating second-order information. Shampoo-style methods use tensor-structured preconditioners, \sophia{} uses lightweight diagonal curvature estimates for language-model pretraining, and SOAP-like methods explore structured preconditioning between these extremes. These approaches promise improved conditioning and token efficiency, but they introduce memory, compute, tuning, and implementation costs. The next section turns to low-rank and projection-based optimizers, which exploit another form of structure in LLM training: the low effective rank of gradients and updates.

\section{Low-Rank and Projection-Based Optimizers}
\label{sec:lowrank}

Low-rank and projection-based optimizers exploit structure in the gradients or updates of large neural networks. Transformer layers contain large dense matrices, and empirical studies suggest that gradients for such matrices often have low effective rank, at least over finite training windows. Projection-based optimizers use this observation to reduce memory by performing optimization in a lower-dimensional subspace while still updating the full parameter matrix. This section reviews the motivation, mechanics, and trade-offs of low-rank optimizer methods, with emphasis on \galore{} and related memory-efficient full-parameter training approaches.

\subsection{Motivation: Low-Rank Structure in Transformer Gradients}
\label{subsec:lowrank_motivation}

Consider a weight matrix $W \in \R^{m \times n}$ in a Transformer block, such as an attention projection or feed-forward projection. A standard optimizer stores gradient and optimizer-state tensors with the same shape as $W$. For \adamw{}, the first- and second-moment states also have shape $m \times n$. This creates a memory cost proportional to $mn$ for each large matrix.

Low-rank methods ask whether the full $m \times n$ update is necessary at every step. If the gradient matrix $G_t = \nabla_W \mathcal{L}_t(W_t)$ has low effective rank, then it may be approximated as
\begin{equation}
    G_t \approx U_t S_t V_t^\top,
\end{equation}
where $U_t \in \R^{m \times r}$, $S_t \in \R^{r \times r}$, $V_t \in \R^{n \times r}$, and $r \ll \min(m,n)$. Instead of storing and optimizing in the full space, one can project the gradient into a lower-dimensional subspace, update optimizer states there, and then project the resulting update back to the original parameter space.

This idea is attractive for LLMs because the largest memory costs are concentrated in large matrices. If low-rank structure can be exploited without substantially degrading optimization, then full-parameter training may become possible with memory closer to parameter-efficient methods, while still updating all model weights.

\subsection{Projection-Based Optimization}
\label{subsec:projection_based_optimization}

A generic projection-based optimizer defines a projection matrix $P_t$ and uses it to map a full gradient into a lower-dimensional representation. For a matrix parameter $W \in \R^{m \times n}$, one simple form is right projection:
\begin{equation}
    \widetilde{G}_t = G_t P_t,
\end{equation}
where $P_t \in \R^{n \times r}$ and $\widetilde{G}_t \in \R^{m \times r}$. The optimizer then maintains states for $\widetilde{G}_t$ rather than $G_t$. After computing an update in the projected space, the update is mapped back:
\begin{equation}
    \Delta W_t = \widetilde{\Delta}_t P_t^\top.
\end{equation}
Other variants may use left projection, two-sided projection, dynamic subspaces, or block-wise projections.

The memory benefit comes from maintaining optimizer states in the projected space. If $r \ll n$, then storing states of shape $m \times r$ is much cheaper than storing states of shape $m \times n$. The main approximation is that updates are restricted to the subspace defined by $P_t$ until the projection is refreshed.

Projection-based optimization therefore introduces a new design question: how should the projection subspace be chosen and updated? A fixed random projection is simple but may miss important gradient directions. A data-dependent projection can better capture gradient structure but requires additional computation. A frequently updated projection adapts quickly but increases overhead; an infrequently updated projection is cheaper but may become stale.

\subsection{GaLore}
\label{subsec:galore}

\galore{} is a representative low-rank optimizer for memory-efficient LLM training~\citep{zhao2024galore}. Its central idea is to project gradients into a low-rank subspace, perform optimizer updates in that subspace, and then project the updates back to the original parameter space. Unlike parameter-efficient fine-tuning methods that freeze most parameters, \galore{} aims to support full-parameter learning with reduced optimizer memory.

At a high level, for a matrix parameter $W$, \galore{} periodically computes a low-rank projection basis from gradient information. The full gradient $G_t$ is projected into a lower-dimensional space, where Adam-style optimizer states can be maintained at reduced cost. If $r$ is the projection rank, the optimizer-state memory for that matrix scales with $r$ rather than the full matrix dimension.

The distinction between \galore{} and low-rank adaptation methods is important. In LoRA-style parameter-efficient fine-tuning, the base weight matrix is frozen and trainable low-rank adapters are added. In \galore{}, the full weight matrix is still updated; low-rank structure is used to reduce the memory cost of the optimizer. Thus, \galore{} is an optimizer/training-memory method rather than a model-architecture adaptation method.

\subsection{Memory Savings and Approximation Error}
\label{subsec:lowrank_memory_approximation}

The memory savings of low-rank projection depend on the rank $r$ and the matrix shape. For a parameter matrix $W \in \R^{m \times n}$, a full Adam-style second-moment state has $mn$ entries. A projected state may have $mr$ or $nr$ entries, depending on projection direction. The memory reduction is substantial when $r \ll \min(m,n)$.

However, projection introduces approximation error. If important gradient components lie outside the selected subspace, the optimizer cannot represent them until the projection is updated. This can slow convergence or bias the training trajectory. The approximation quality depends on the effective rank of gradients, the projection update frequency, and the stability of gradient subspaces over time.

This creates a trade-off:
\begin{equation}
    \text{lower rank} \Rightarrow \text{more memory savings but more approximation error},
\end{equation}
while
\begin{equation}
    \text{higher rank} \Rightarrow \text{less approximation error but less memory savings}.
\end{equation}
Choosing the rank is therefore analogous to choosing a compression level.

\subsection{Projection Refresh and Subspace Staleness}
\label{subsec:projection_refresh}

Projection-based methods often refresh the low-rank subspace periodically rather than at every step. Let $K$ denote the projection refresh interval. If $K=1$, the projection adapts every step, but the overhead may be high. If $K$ is large, the projection is cheaper but may become stale as the optimization trajectory changes.

Subspace staleness is especially relevant in LLM training because gradient statistics can change across phases. Early training, mid-training, and late training may have different dominant directions. Fine-tuning may also produce rapidly changing gradients when the downstream dataset is small or distribution-shifted. Projection refresh policies must therefore balance stability and adaptivity.

A useful evaluation should report projection rank, refresh frequency, projection computation cost, and how these choices affect both memory and convergence. Without these details, it is difficult to compare low-rank optimizers with full-state baselines.

\subsection{Relationship to Parameter-Efficient Fine-Tuning}
\label{subsec:lowrank_vs_peft}

Low-rank optimizers are related to but distinct from parameter-efficient fine-tuning methods. Parameter-efficient methods reduce training memory by freezing most model parameters and training a small set of added or selected parameters. Low-rank optimizers, in contrast, reduce optimizer-state memory while still updating the original parameters.

This distinction matters for both capability and evaluation. Parameter-efficient fine-tuning may be sufficient for many adaptation tasks, but it may not match full-parameter fine-tuning in all regimes. Low-rank optimizers aim to preserve the expressivity of full-parameter updates while reducing memory. They should therefore be compared not only with \adamw{} full fine-tuning, but also with parameter-efficient baselines when the target application is adaptation under limited memory.

In pretraining, the distinction is even clearer. Parameter-efficient adapters are not usually used to train a base model from scratch, whereas low-rank optimizers can in principle support full-model pretraining. This makes projection-based methods part of the optimizer landscape rather than merely the fine-tuning landscape.

\subsection{APOLLO-like Projection and Low-Memory Adaptivity}
\label{subsec:apollo_projection}

APOLLO-like methods aim to achieve \adamw{}-level performance with memory closer to \sgd{}~\citep{han2024apollo}. While different methods use different mechanisms, they share the goal of avoiding full per-parameter adaptive state while preserving effective optimization. Some approaches combine projection, structured statistics, or approximate adaptive scaling.

These methods highlight a continuum between memory-efficient adaptive optimizers and low-rank projection methods. At one end, Adam-mini reduces redundant coordinate-wise adaptivity through grouping. At another, \galore{} explicitly projects gradients into a low-rank subspace. APOLLO-like approaches occupy the broader space of low-memory adaptive training, where the optimizer attempts to retain the useful behavior of \adamw{} without storing all of its state.

Because this area is developing quickly, careful reporting is essential. Claims of \adamw{}-level performance should specify model scale, dataset, token budget, tuning budget, precision, and memory accounting. The central question is whether the method remains competitive under strong baselines and realistic LLM training regimes.

\subsection{Low-Rank Methods and Distributed Training}
\label{subsec:lowrank_distributed}

Low-rank optimizers interact with distributed training in several ways. On the positive side, reduced optimizer-state memory can lower per-device memory requirements and may reduce checkpoint size. Projection may also reduce the memory footprint of optimizer states under sharding.

On the negative side, projection bases must be computed, stored, and sometimes synchronized. If the projection depends on global gradient information, distributed implementations must coordinate across workers. If projection is applied independently across shards, the resulting subspaces may differ from those computed on the full matrix. These details can affect both correctness and performance.

For tensor-parallel LLM training, large weight matrices may be partitioned across devices. A projection computed for a shard may not capture the same low-rank structure as a projection computed for the full matrix. This raises implementation questions that are not visible in single-device algorithm descriptions.

\subsection{Evaluation Criteria}
\label{subsec:lowrank_evaluation}

Low-rank and projection-based optimizers should be evaluated using both optimization and compression metrics. Important quantities include:
\begin{enumerate}[leftmargin=*]
    \item projection rank $r$;
    \item projection refresh interval $K$;
    \item memory used by optimizer states and projection bases;
    \item cost of computing projections;
    \item validation loss at fixed token budget;
    \item wall-clock time to target loss;
    \item downstream task performance;
    \item sensitivity to rank and refresh frequency;
    \item compatibility with sharding and tensor parallelism.
\end{enumerate}

A strong evaluation should include rank-memory-convergence trade-off curves. These curves reveal whether memory savings degrade gracefully or whether performance collapses below a critical rank. They also help distinguish robust low-rank structure from brittle compression.

\subsection{Strengths and Limitations}
\label{subsec:lowrank_strengths_limitations}

\paragraph{Strengths.}
Low-rank optimizers can reduce optimizer-state memory while preserving full-parameter updates. They exploit the matrix-heavy structure of Transformers and can be especially useful when large dense matrices dominate memory. They also provide a bridge between adaptive optimization and model-structure-aware training.

\paragraph{Limitations.}
The main limitation is approximation error. Projection restricts updates to a subspace, and important directions may be missed. Projection computation introduces overhead, and distributed implementations can be complex. The method also introduces new hyperparameters, such as rank and refresh interval.

\paragraph{Open empirical question.}
The key empirical question is whether low-rank gradient structure is stable and strong enough across LLM scales, datasets, and training phases to justify replacing full-state optimizers. Evidence from methods such as \galore{} is promising, but broader benchmarking is needed.

\subsection{Summary}
\label{subsec:lowrank_summary}

Low-rank and projection-based optimizers reduce memory by exploiting structure in gradient or update matrices. \galore{} is a central example: it performs optimization in a low-rank projected space while still updating full model parameters. These methods differ from parameter-efficient fine-tuning because they do not freeze the base model. Their value depends on the trade-off among memory savings, projection error, refresh overhead, and distributed-training compatibility. The next section examines matrix-based and orthogonalized optimizers, which also exploit matrix structure but modify update geometry through matrix normalization or orthogonalization rather than low-rank projection.

\section{Matrix-Based and Orthogonalized Optimizers}
\label{sec:matrix}

Matrix-based optimizers exploit the fact that most parameters in Transformer language models are not arbitrary vectors, but large matrices with meaningful input--output structure. Attention projections, feed-forward projections, and output layers are all matrix-valued parameters whose geometry may matter for optimization. Standard optimizers such as \adamw{} treat these matrices as collections of independent scalar coordinates. Matrix-based optimizers instead modify the update at the matrix level, often through normalization, orthogonalization, or spectral control. This section reviews the motivation for matrix-aware optimization, with emphasis on \muon{} and related orthogonalized update methods.

\subsection{Why Matrix Geometry Matters}
\label{subsec:matrix_geometry_motivation}

A Transformer block is dominated by matrix multiplications. For an input activation matrix $X$, a linear projection computes
\begin{equation}
    Y = XW,
\end{equation}
where $W \in \R^{d_{\mathrm{in}} \times d_{\mathrm{out}}}$ is a weight matrix. During training, the optimizer updates $W$ using a gradient matrix $G_t = \nabla_W \mathcal{L}_t(W_t)$. A coordinate-wise optimizer treats $G_t$ as a set of independent entries, while a matrix-aware optimizer treats it as a structured linear map.

This distinction matters because the singular values, rank, and orientation of an update matrix can affect how the layer transforms activations. A coordinate-wise step may have undesirable spectral properties even if each individual coordinate update is well scaled. Matrix-based optimizers attempt to control update geometry at the level of the full matrix rather than at the level of individual entries.

The matrix perspective is different from the low-rank perspective in \Cref{sec:lowrank}. Low-rank optimizers compress updates into a lower-dimensional subspace to save memory. Matrix-based optimizers may use full-rank updates, but transform them to have desirable matrix-level properties such as approximate orthogonality or controlled spectral norm.

\subsection{From Vector Updates to Matrix Updates}
\label{subsec:vector_to_matrix_updates}

A standard optimizer can be written as
\begin{equation}
    W_{t+1} = W_t - \eta_t U_t,
\end{equation}
where $U_t$ is an update matrix derived from gradients and optimizer states. In \adamw{}, $U_t$ is obtained by applying coordinate-wise scaling to the momentum estimate. In a matrix-based optimizer, $U_t$ may be transformed by a matrix function:
\begin{equation}
    W_{t+1} = W_t - \eta_t \Phi(U_t),
\end{equation}
where $\Phi$ modifies the update according to matrix-level criteria.

Examples of matrix-level transformations include normalization by Frobenius norm, normalization by spectral norm, orthogonalization, projection onto a manifold, or approximate whitening. The goal is to produce an update whose action as a linear map is better conditioned than the raw gradient or momentum update.

This formulation highlights the central design question: what matrix function $\Phi$ improves LLM training while remaining cheap enough for large-scale use?

\subsection{Orthogonalized Updates}
\label{subsec:orthogonalized_updates}

One approach is to orthogonalize the update matrix. For a matrix $U \in \R^{m \times n}$, an orthogonalized version can be obtained from its polar decomposition. If $U$ has singular value decomposition
\begin{equation}
    U = A \Sigma B^\top,
\end{equation}
then a matrix with the same singular vectors but unit singular values is
\begin{equation}
    \operatorname{orth}(U) = A B^\top.
\end{equation}
This transformation preserves the left and right singular directions of $U$ but normalizes its singular values.

Computing a full singular value decomposition at every optimizer step is too expensive for LLM training. Practical methods therefore use iterative approximations, such as Newton--Schulz iterations, to approximate orthogonalization. The resulting update controls matrix geometry without requiring exact decompositions.

Orthogonalized updates are motivated by the idea that an update matrix with balanced singular values may produce more stable and effective changes to a layer than one dominated by a few large singular directions. This is related to, but distinct from, adaptive diagonal scaling. Diagonal adaptivity equalizes coordinate scales, while orthogonalization equalizes singular-value structure.

\subsection{Muon}
\label{subsec:muon}

\muon{} is a recent matrix-based optimizer that applies orthogonalization to momentum updates for neural-network weight matrices~\citep{jordan2024muon}. The name is commonly interpreted as MomentUm Orthogonalized by Newton--Schulz. At a high level, \muon{} first forms a momentum update and then applies an approximate matrix orthogonalization procedure before updating the parameter matrix.

A simplified schematic update is
\begin{align}
    M_t &= \beta M_{t-1} + G_t, \\
    U_t &= \operatorname{NS}(M_t), \\
    W_{t+1} &= W_t - \eta_t U_t,
\end{align}
where $M_t$ is a momentum buffer and $\operatorname{NS}(\cdot)$ denotes an approximate Newton--Schulz orthogonalization operator. This schematic omits implementation details such as normalization, rectangular matrix handling, weight decay, and parameter-group choices, but captures the defining feature: the update is a matrix-normalized momentum direction.

\muon{} differs from \adamw{} in two major ways. First, it does not rely on a coordinate-wise second-moment accumulator. Second, it transforms updates according to matrix geometry. Compared with \lion{}, which also avoids second moments but uses coordinate-wise signs, \muon{} uses a matrix-level normalization. Compared with \galore{}, which reduces memory through low-rank projection, \muon{} focuses on the geometry of full matrix updates.

Recent work has investigated the scalability of \muon{} for LLM training~\citep{liu2025muon}. These studies are important because matrix-based update rules may behave differently at small scale and LLM scale. A method that improves small models may not necessarily retain its advantage when applied to billion-parameter Transformers under realistic training budgets.

\subsection{Newton--Schulz Orthogonalization}
\label{subsec:newton_schulz}

Newton--Schulz iteration is an iterative method that can approximate matrix inverse roots or polar factors. For optimizer purposes, the goal is to transform an update matrix into a more orthogonal or spectrally normalized form without computing a full singular value decomposition.

A simplified Newton--Schulz-style iteration for polar decomposition starts with a normalized matrix $X_0$ and repeatedly applies a polynomial iteration such as
\begin{equation}
    X_{k+1} = a X_k + b X_k X_k^\top X_k + c X_k X_k^\top X_k X_k^\top X_k,
\end{equation}
with coefficients chosen to drive the singular values toward one. Practical implementations vary in the exact polynomial, normalization, number of iterations, and numerical precision.

The key advantage of Newton--Schulz-style orthogonalization is that it uses matrix multiplications, which are efficient on modern accelerators. The key cost is additional matrix multiplication per optimizer step. For LLMs, this overhead must be compared against potential reductions in training tokens or improvements in stability.

\subsection{Which Parameters Should Use Matrix-Based Updates?}
\label{subsec:which_params_matrix}

Not all parameters in an LLM are equally suitable for matrix-based optimization. Large dense matrices are natural candidates, including:
\begin{enumerate}[leftmargin=*]
    \item attention query, key, value, and output projections;
    \item feed-forward up-projection, down-projection, and gated projection matrices;
    \item large output or embedding matrices, depending on implementation;
    \item other dense linear layers in multimodal or mixture-of-experts variants.
\end{enumerate}

Biases, normalization parameters, scalar parameters, and small tensors may not benefit from matrix orthogonalization. Practical matrix-based optimizers therefore often use hybrid parameter grouping: matrix parameters receive matrix-based updates, while other parameters use \adamw{}, \sgd{}, or another simpler optimizer.

This hybrid nature complicates evaluation. A reported result may depend not only on the matrix optimizer itself, but also on which parameters are assigned to it and which optimizer is used for the remaining parameters. Therefore, papers should report parameter-group rules explicitly.

\subsection{Relationship to AdamW, Lion, and GaLore}
\label{subsec:matrix_relationships}

Matrix-based optimizers are best understood by contrasting them with nearby families.

\paragraph{Compared with AdamW.}
\adamw{} uses coordinate-wise first- and second-moment estimates. It adapts each scalar parameter independently. Matrix-based optimizers instead normalize or transform an entire update matrix. This can exploit correlations among coordinates but may lose the fine-grained coordinate-wise scaling of \adamw{}.

\paragraph{Compared with Lion.}
\lion{} applies a sign operation to a momentum-like vector. This is a coordinate-wise normalization. \muon{} applies matrix-level orthogonalization to a momentum-like matrix. Both simplify or replace second-moment adaptivity, but they impose different geometries.

\paragraph{Compared with GaLore.}
\galore{} uses low-rank projection to reduce optimizer-state memory while retaining full-parameter updates. \muon{} changes the shape of the update through orthogonalization. The former is primarily a memory-reduction strategy; the latter is primarily an update-geometry strategy, although it may also reduce state relative to \adamw{} by avoiding second moments.

\paragraph{Compared with Shampoo.}
Shampoo-style methods use matrix or tensor preconditioners derived from gradient covariance statistics. \muon{} uses matrix normalization or orthogonalization of the update itself. Both are matrix-aware, but Shampoo is closer to curvature-aware preconditioning, while \muon{} is closer to normalized momentum.

\subsection{Memory, Compute, and Implementation Trade-Offs}
\label{subsec:matrix_tradeoffs}

Matrix-based optimizers have a distinctive trade-off profile.

\paragraph{Memory.}
If a matrix-based optimizer stores only momentum and avoids second moments, it can reduce optimizer-state memory relative to \adamw{}. However, it may require temporary buffers for orthogonalization. The net memory benefit depends on implementation.

\paragraph{Compute.}
Orthogonalization requires additional matrix operations. Newton--Schulz iterations can be efficient on accelerators, but they still add work relative to simple element-wise updates. The relevant metric is whether the improved update geometry reduces the number of tokens or wall-clock time needed to reach a target loss.

\paragraph{Numerical stability.}
Matrix normalization can be sensitive to scaling, precision, and singular-value distributions. Practical implementations must normalize inputs to the iteration and choose iteration counts carefully.

\paragraph{Parameter grouping.}
Hybrid optimization rules introduce additional design choices. Which matrices should use \muon{}? Which parameters should use \adamw{}? Should embeddings and output heads be treated differently? These choices affect results and should be reported.

\paragraph{Distributed training.}
When matrices are partitioned across tensor-parallel devices, orthogonalizing a shard may differ from orthogonalizing the full matrix. Distributed implementations must decide whether to operate locally on shards or globally across partitions. This can affect both efficiency and update geometry.

\subsection{Theoretical Questions}
\label{subsec:matrix_theory}

Matrix-based optimizers raise theoretical questions that are not fully answered.

\paragraph{What objective geometry is being approximated?}
Diagonal adaptive methods can be interpreted as coordinate-wise preconditioners, while Newton-style methods approximate curvature. The theoretical interpretation of orthogonalized momentum updates is less settled. One question is whether they approximate steepest descent under a matrix norm or impose beneficial spectral regularization on updates.

\paragraph{Why should singular-value normalization help?}
Orthogonalized updates equalize singular values of the update matrix. This may prevent updates from being dominated by a small number of directions, but it may also discard useful magnitude information. Understanding when this helps is an open problem.

\paragraph{How does matrix geometry interact with Transformer training dynamics?}
Attention and MLP matrices play different roles in the model. Matrix-based updates may affect representation learning, feature rotation, and layer conditioning differently across these components.

\paragraph{How should rectangular matrices be handled?}
Transformer matrices are often rectangular, especially in feed-forward layers. Orthogonalization behavior can depend on whether the matrix is tall or wide and whether the update is normalized along input or output dimensions.

Recent theoretical work has begun to analyze convergence properties of \muon{}-style Newton--Schulz updates~\citep{kim2026convergence}, but a complete theory of matrix-based optimization for LLMs remains open.

\subsection{Evaluation Guidelines}
\label{subsec:matrix_evaluation}

Matrix-based optimizers should be evaluated with implementation details that are usually
irrelevant for purely element-wise optimizers. In addition to the general protocol in
\Cref{sec:benchmarking}, papers should report which parameter groups use matrix-based
updates, which optimizer is used for non-matrix parameters, the number of Newton--Schulz
or orthogonalization iterations, normalization rules, temporary-buffer memory, additional
matrix-multiplication cost, and behavior under tensor parallelism or sharding.

\subsection{Summary}
\label{subsec:matrix_summary}

Matrix-based optimizers exploit the fact that Transformer parameters are dominated by large matrices. Rather than treating each coordinate independently, they transform updates according to matrix-level geometry. \muon{} is a representative method that orthogonalizes momentum updates using Newton--Schulz-style iterations, aiming to produce well-conditioned matrix updates without full second-moment adaptivity. This family is conceptually distinct from \adamw{}, \lion{}, \galore{}, and Shampoo: it replaces coordinate-wise or low-rank structure with matrix-level update normalization. Its promise lies in better exploiting Transformer architecture, while its challenges include computational overhead, parameter grouping, distributed implementation, and theoretical understanding. The next section turns from optimizer families to benchmarking methodology, asking how such diverse optimizers should be compared fairly.

\section{Benchmarking Optimizers for LLMs}
\label{sec:benchmarking}

Optimizer benchmarking for LLMs is difficult because optimizers are embedded in complex training systems. A reported improvement may arise from a better update rule, a better-tuned baseline, a more favorable learning-rate schedule, a different batch size, a faster implementation, or a scale regime where one method happens to be advantaged. This section discusses how LLM optimizers should be compared and what information is needed to interpret empirical claims.

\subsection{What Does It Mean for an Optimizer to Be Better?}
\label{subsec:what_better_optimizer}

There is no single universal criterion for optimizer quality. An optimizer may be better according to one metric and worse according to another. For LLM training, the most important criteria include:
\begin{enumerate}[leftmargin=*]
    \item \textbf{Token efficiency:} validation loss or downstream performance as a function of training tokens.
    \item \textbf{Step efficiency:} progress per optimizer step.
    \item \textbf{Wall-clock efficiency:} time to reach a target validation loss.
    \item \textbf{Memory efficiency:} peak memory usage and optimizer-state size.
    \item \textbf{Compute efficiency:} floating-point operations or accelerator hours required to reach a target loss.
    \item \textbf{Stability:} robustness to seeds, learning rates, batch sizes, and precision settings.
    \item \textbf{Scalability:} whether gains persist as model size, data size, context length, and batch size increase.
    \item \textbf{Implementation complexity:} compatibility with fused kernels, sharding, checkpointing, and distributed training systems.
\end{enumerate}

These criteria can conflict. A curvature-aware optimizer may improve token efficiency but increase per-step cost. A memory-efficient optimizer may slightly reduce convergence speed but allow a larger model under the same hardware budget. A sign-based optimizer may be simple and memory efficient but require careful learning-rate tuning. Thus, optimizer benchmarking should be treated as a multi-objective evaluation problem rather than a single leaderboard.

\subsection{Fixed-Model Versus Fixed-Resource Evaluation}
\label{subsec:fixed_model_fixed_resource}

Two evaluation regimes are especially important.

\paragraph{Fixed-model evaluation.}
In fixed-model evaluation, all optimizers train the same model architecture with the same dataset, token budget, batch size, precision, and hardware. This isolates the optimizer as much as possible. It answers the question: \emph{given this exact training setup, which optimizer performs best?}

\paragraph{Fixed-resource evaluation.}
In fixed-resource evaluation, all optimizers are given the same hardware, memory budget, wall-clock budget, or compute budget, but may use different feasible configurations. This answers a different question: \emph{given the same resources, which optimizer enables the best final model?}

Both regimes are necessary. Fixed-model comparisons are cleaner scientifically, but fixed-resource comparisons may be more relevant practically. For example, a memory-efficient optimizer may not outperform \adamw{} on the same model, but may enable a larger model or longer context that achieves better final performance under the same memory budget.

\subsection{The Baseline Problem}
\label{subsec:baseline_problem}

Most new LLM optimizers compare against \adamw{}. This is appropriate because \adamw{} is the dominant practical baseline. However, the comparison is only meaningful if the \adamw{} baseline is strong. A weak baseline can make a new optimizer appear better than it is.

A strong \adamw{} baseline should include appropriate tuning of:
\begin{enumerate}[leftmargin=*]
    \item learning rate;
    \item warmup length;
    \item decay schedule;
    \item $\beta_1$ and $\beta_2$;
    \item weight decay;
    \item gradient clipping;
    \item batch size and gradient accumulation;
    \item parameter-specific weight-decay exclusions;
    \item precision and loss-scaling settings.
\end{enumerate}

The tuning budget should be comparable across optimizers. If a new optimizer receives extensive tuning but \adamw{} uses default hyperparameters, the comparison is biased. Conversely, applying \adamw{} hyperparameters to a new optimizer may unfairly disadvantage the new method. Recent empirical studies emphasize that optimizer conclusions can change substantially under fairer tuning protocols~\citep{zhao2025deconstructing,semenov2025benchmarking}.

\subsection{Hyperparameter Search and Reporting}
\label{subsec:hyperparameter_reporting}

Optimizer papers should report the hyperparameter search space and selection procedure. At minimum, they should specify:
\begin{enumerate}[leftmargin=*]
    \item values tried for learning rate and weight decay;
    \item schedule type and warmup fraction;
    \item momentum or decay coefficients;
    \item optimizer-specific hyperparameters;
    \item number of trials per optimizer;
    \item selection criterion used to choose the final configuration;
    \item whether hyperparameters were tuned on validation loss, downstream tasks, or training loss;
    \item whether the same tuning budget was used for all baselines.
\end{enumerate}

Reporting only the best result without search details makes it difficult to determine whether improvements are robust. For expensive LLM pretraining, exhaustive tuning may be infeasible, but limited tuning should be stated clearly. A useful practice is to report sensitivity curves for the most important hyperparameters, especially learning rate and weight decay.

\subsection{Scale Dependence}
\label{subsec:benchmark_scale_dependence}

Optimizer behavior can change with scale. A method that improves a 100-million-parameter model may not improve a 7-billion-parameter model. Similarly, a method that helps during the first few billion tokens may not improve final loss after hundreds of billions of tokens.

Scale dependence arises from several sources:
\begin{enumerate}[leftmargin=*]
    \item larger models may have different gradient noise scales;
    \item deeper networks may be more sensitive to update geometry;
    \item larger datasets change the relative importance of early and late training;
    \item larger batch sizes change gradient variance;
    \item distributed systems change the cost of optimizer states and communication;
    \item longer context lengths change memory pressure and activation costs.
\end{enumerate}

Therefore, optimizer evaluations should include multiple model scales when possible. If large-scale experiments are infeasible, papers should avoid overclaiming from small-scale results. Scaling trends are often more informative than a single endpoint.

\subsection{Early-Training Curves and Final Performance}
\label{subsec:early_vs_final}

Early training curves can be misleading. An optimizer may reduce loss quickly in the initial phase but converge to a worse final loss. Another optimizer may appear slower early but achieve better final performance after a longer schedule. This is especially important for methods with different implicit regularization or update geometry.

A robust evaluation should report:
\begin{enumerate}[leftmargin=*]
    \item early validation curves;
    \item final validation loss after the full token budget;
    \item downstream task performance at final checkpoint;
    \item intermediate checkpoints if the optimizer is claimed to improve sample efficiency;
    \item whether learning-rate schedules were adjusted for the full training horizon.
\end{enumerate}

Claims about speedup should specify the target loss and the training interval over which the speedup is measured. A method that reaches an intermediate loss faster does not necessarily reduce total training cost if the advantage disappears later.

\subsection{Token Efficiency Versus Wall-Clock Efficiency}
\label{subsec:token_vs_wallclock}

Token efficiency measures progress as a function of the number of training tokens. Wall-clock efficiency measures progress as a function of time. Both are important, and they can disagree.

Curvature-aware methods may improve token efficiency but have higher per-step cost. Low-rank projection methods may reduce memory but require projection updates. Matrix-based optimizers may use additional matrix multiplications for orthogonalization. Quantized optimizers may reduce memory but introduce quantization and dequantization overhead. In each case, validation loss versus tokens is insufficient.

A complete benchmark should report validation loss as a function of both tokens and wall-clock time. When wall-clock measurements are unavailable, papers should report per-step overhead, additional FLOPs, memory bandwidth costs, and implementation details that allow approximate comparison.

\subsection{Memory Accounting}
\label{subsec:benchmark_memory_accounting}

Memory accounting is central for LLM optimizer evaluation. Papers should distinguish among:
\begin{enumerate}[leftmargin=*]
    \item model parameter memory;
    \item gradient memory;
    \item optimizer-state memory;
    \item activation memory;
    \item temporary buffers;
    \item communication buffers;
    \item checkpoint memory;
    \item memory after sharding or partitioning.
\end{enumerate}

Optimizer-state memory alone can be misleading. For example, an optimizer may reduce moment-state memory but increase temporary buffer memory. Another may reduce per-device memory only when combined with sharding. Memory should therefore be reported as both theoretical state size and measured peak memory in a concrete training run.

For memory-efficient optimizers, two comparisons are useful. The first is fixed-model memory reduction: how much memory is saved on the same model? The second is fixed-resource improvement: what larger model, longer context, or larger batch becomes feasible under the same hardware budget?

\subsection{Precision, Numerics, and Stability}
\label{subsec:benchmark_precision}

LLM training commonly uses mixed precision. Optimizer comparisons should report the precision used for:
\begin{enumerate}[leftmargin=*]
    \item model weights;
    \item gradients;
    \item optimizer states;
    \item master weights;
    \item preconditioners or curvature estimates;
    \item matrix orthogonalization or projection computations.
\end{enumerate}

Precision choices can strongly affect stability. For example, curvature estimates may require higher precision than model weights. Quantized optimizer states may behave differently across hardware. Matrix orthogonalization may be sensitive to normalization and numerical range. Gradient clipping, loss scaling, and epsilon values should also be reported.

Stability should be evaluated across random seeds when feasible. At minimum, papers should report whether any runs diverged, whether gradient clipping was necessary, and whether the optimizer was sensitive to small hyperparameter changes.

\subsection{Downstream Evaluation}
\label{subsec:benchmark_downstream}

Validation loss is the primary metric for pretraining optimization, but downstream performance is also important. Two optimizers may reach similar validation loss while producing models with different downstream behavior. This can occur because optimizers may have different implicit regularization effects or may traverse different training trajectories.

Downstream evaluation should be interpreted carefully. Benchmark variance, prompt formatting, evaluation contamination, and decoding settings can all affect results. For optimizer comparisons, downstream tasks should supplement validation loss rather than replace it. A good evaluation reports both language-modeling loss and downstream performance under standardized evaluation settings.

\subsection{Implementation Effects}
\label{subsec:benchmark_implementation}

The practical speed of an optimizer depends heavily on implementation. \adamw{} benefits from highly optimized fused kernels and mature distributed support. New optimizers may be slower simply because their implementations are less optimized, or faster because they use specialized kernels not available to baselines.

Implementation details to report include:
\begin{enumerate}[leftmargin=*]
    \item framework and version;
    \item accelerator type;
    \item fused or unfused optimizer kernels;
    \item distributed-training strategy;
    \item sharding configuration;
    \item checkpointing strategy;
    \item whether custom kernels are used;
    \item measured tokens per second.
\end{enumerate}

A mathematically superior optimizer may fail to improve wall-clock time if implementation overhead is high. Conversely, a well-engineered implementation can make a modest algorithmic improvement practically valuable.

\subsection{Recommended Benchmark Protocol}
\label{subsec:recommended_protocol}

Based on the preceding discussion, we recommend that LLM optimizer papers report results under a protocol with the following components.

\paragraph{Model and data.}
Specify architecture, parameter count, tokenizer, dataset, sequence length, and number of training tokens.

\paragraph{Training setup.}
Specify global batch size in tokens, microbatch size, gradient accumulation, precision, hardware, and distributed configuration.

\paragraph{Optimizer configuration.}
Specify update equations, state precision, hyperparameters, learning-rate schedule, warmup, weight decay, clipping, and parameter-group rules.

\paragraph{Tuning protocol.}
Report search spaces, tuning budgets, selection criteria, and whether baselines received comparable tuning.

\paragraph{Metrics.}
Report validation loss versus tokens, validation loss versus wall-clock time, peak memory, tokens per second, downstream performance, and run stability.

\paragraph{Ablations.}
Include ablations for optimizer-specific components, such as projection rank, curvature update frequency, quantization precision, or orthogonalization iterations.

\paragraph{Scale.}
Evaluate at multiple model sizes or token budgets when possible. If only small-scale experiments are feasible, state the limitation explicitly.

\subsection{Common Pitfalls}
\label{subsec:benchmark_pitfalls}

Several pitfalls recur in optimizer benchmarking.

\paragraph{Under-tuned baselines.}
Comparing against default \adamw{} settings can exaggerate improvements.

\paragraph{Unequal tuning budgets.}
Giving the proposed optimizer more search effort than baselines biases results.

\paragraph{Early-curve overclaiming.}
Reporting speedups only during early training can be misleading if gains vanish later.

\paragraph{Ignoring wall-clock cost.}
Token-efficiency improvements may not matter if per-step overhead is too large.

\paragraph{Incomplete memory reporting.}
Reporting only optimizer-state memory can hide activation, temporary-buffer, or sharding effects.

\paragraph{Small-scale extrapolation.}
Results on small models may not predict behavior at LLM scale.

\paragraph{Implementation confounding.}
Differences in kernel quality, sharding support, or framework integration can dominate algorithmic differences.

\subsection{Summary}
\label{subsec:benchmarking_summary}

Benchmarking LLM optimizers requires a multi-objective view. The relevant question is not simply which optimizer achieves the lowest validation loss in a single experiment, but which optimizer provides the best trade-off among token efficiency, wall-clock time, memory, stability, scalability, and implementation complexity. Fair comparison requires strong baselines, transparent tuning, careful memory accounting, and scale-aware evaluation. The next section turns from evaluation methodology to open problems, highlighting directions where optimizer research for LLMs remains incomplete.

\section{Open Problems and Future Directions}
\label{sec:openproblems}

Optimizer research for LLMs is moving quickly, but many fundamental questions remain unresolved. The field lacks reliable scaling laws for optimizer behavior, standardized evaluation protocols, strong theory for matrix-based updates, and clear guidance on when memory-efficient or curvature-aware methods improve end-to-end training. This section summarizes open problems that are especially important for future work.

\subsection{Scaling Laws for Optimizers}
\label{subsec:open_scaling_laws}

LLM scaling laws have clarified how model size, dataset size, and compute budget affect loss, but comparable scaling laws for optimizers remain underdeveloped. In practice, optimizers can behave differently across model sizes, token budgets, and batch sizes. An optimizer that appears superior at small scale may lose its advantage at larger scale, while another may require large models or long training horizons before its benefits appear.

A useful optimizer scaling law would answer questions such as:
\begin{enumerate}[leftmargin=*]
    \item How does the optimal learning rate vary with model size, batch size, and token budget for each optimizer?
    \item Do memory-efficient optimizers preserve their convergence behavior as model size increases?
    \item Do curvature-aware or matrix-based optimizers become more or less useful at larger scale?
    \item How does the critical batch size depend on the optimizer?
    \item Can small-scale experiments reliably predict large-scale optimizer rankings?
\end{enumerate}

Recent benchmarking studies suggest that optimizer conclusions can be sensitive to scale and tuning~\citep{zhao2025deconstructing,semenov2025benchmarking}. Developing predictive scaling rules would make optimizer research less dependent on expensive full-scale training runs.

\subsection{Fair and Reproducible Optimizer Benchmarks}
\label{subsec:open_fair_benchmarks}

The field needs standardized optimizer benchmarks for LLM training. Existing comparisons often differ in architecture, dataset, hyperparameter budget, implementation quality, and reporting detail. This makes it difficult to determine whether a method improves optimization or merely benefits from a favorable experimental setup.

A strong benchmark suite should include:
\begin{enumerate}[leftmargin=*]
    \item multiple model sizes, from small proxies to billion-parameter models;
    \item standardized datasets and token budgets;
    \item fixed-model and fixed-resource evaluation tracks;
    \item transparent hyperparameter search spaces;
    \item wall-clock, memory, and validation-loss reporting;
    \item downstream evaluation under standardized settings;
    \item reproducible training code and optimizer implementations.
\end{enumerate}

Such benchmarks should also include strong \adamw{} baselines. Since \adamw{} is highly optimized and widely tuned, new methods should be compared against competitive implementations rather than default settings. Benchmarking should also separate algorithmic gains from systems gains whenever possible.

\subsection{Memory--Compute--Convergence Trade-Offs}
\label{subsec:open_memory_compute_convergence}

Many new optimizers improve one resource at the expense of another. Memory-efficient optimizers reduce state size but may require more training tokens. Curvature-aware optimizers can reduce the number of steps but increase per-step computation. Matrix-based optimizers may improve update geometry but add orthogonalization overhead. Low-rank methods reduce state memory but introduce projection error and refresh costs.

The open problem is to characterize these trade-offs quantitatively. A useful comparison should answer:
\begin{equation}
    \text{Which optimizer minimizes final loss under a fixed hardware, memory, and time budget?}
\end{equation}
This is different from asking which optimizer has the best loss per token or the smallest state size in isolation.

Future work should evaluate optimizers using joint resource frontiers. For example, one can plot final validation loss against peak memory, wall-clock time, or total accelerator hours. Such Pareto frontiers would make trade-offs explicit and help practitioners choose optimizers for different regimes.

\subsection{Optimizer-State Memory Beyond AdamW}
\label{subsec:open_state_memory}

A central question is how much optimizer state is truly necessary for LLM training. \adamw{} stores two full moment vectors, but methods such as \adafactor{}, Adam-mini, \lomo{}, \galore{}, and APOLLO-like approaches suggest that full per-coordinate adaptivity may not always be required~\citep{shazeer2018adafactor,zhang2025adammini,lv2024lomo,zhao2024galore,han2024apollo}.

Several questions remain open:
\begin{enumerate}[leftmargin=*]
    \item Which parameters actually benefit from full second-moment adaptivity?
    \item Can different layers or parameter groups use different levels of optimizer state?
    \item Is second-moment memory more important early or late in training?
    \item Can optimizer states be compressed dynamically as training progresses?
    \item How do memory-efficient states interact with sharding, checkpointing, and mixed precision?
\end{enumerate}

A promising direction is adaptive state allocation: use rich optimizer states only where they matter most, and cheaper updates elsewhere. This would treat optimizer memory as a budget to allocate across the model rather than a fixed multiple of parameter count.

\subsection{Parameter-Group-Specific Optimizers}
\label{subsec:open_parameter_group_optimizers}

Modern LLMs contain heterogeneous parameter groups. Attention projections, feed-forward matrices, embeddings, normalization parameters, and output heads may have different gradient statistics and optimization needs. Yet most training recipes apply a single optimizer to nearly all parameters, with limited exceptions such as excluding normalization parameters from weight decay.

Future optimizers may be hybrid by design. For example, one could use a matrix-based optimizer for large dense matrices, \adamw{} or Adam-mini for embeddings, \sgd{} or momentum for normalization parameters, and low-rank projection for selected feed-forward layers. The challenge is to design such hybrids systematically rather than through ad hoc parameter grouping.

Open questions include:
\begin{enumerate}[leftmargin=*]
    \item Which parameter groups require coordinate-wise adaptivity?
    \item Which matrices benefit from orthogonalized or matrix-normalized updates?
    \item Should embeddings and output heads use different optimizers from internal layers?
    \item Can optimizer assignment be learned or adapted during training?
    \item How should weight decay and learning-rate schedules vary by parameter group?
\end{enumerate}

This direction connects optimizer design with architecture-aware training. It may be especially important for mixture-of-experts models, multimodal LLMs, and long-context architectures.

\subsection{Theory for Matrix-Based Optimizers}
\label{subsec:open_matrix_theory}

Matrix-based optimizers such as \muon{} raise theoretical questions that are not fully addressed by existing analyses of diagonal adaptive methods or classical second-order optimization~\citep{jordan2024muon,liu2025muon,kim2026convergence}. These methods transform update matrices using orthogonalization or matrix normalization, but the precise optimization principle behind these transformations remains an active area of study.

Important theoretical questions include:
\begin{enumerate}[leftmargin=*]
    \item Under what norm or geometry is an orthogonalized update a natural descent direction?
    \item When does singular-value normalization improve conditioning, and when does it discard useful magnitude information?
    \item How do matrix-normalized updates interact with Transformer activations and residual connections?
    \item What convergence guarantees are possible for Newton--Schulz-style optimizer updates?
    \item How should rectangular matrices and tensor-parallel shards be treated theoretically?
\end{enumerate}

A stronger theory would help distinguish principled matrix-aware optimization from empirical update normalization. It could also guide which parameter groups should receive matrix-based updates.

\subsection{Curvature Without Prohibitive Overhead}
\label{subsec:open_curvature_overhead}

Second-order information is attractive because it can improve conditioning, but exact curvature is impossible at LLM scale. Methods such as Shampoo and \sophia{} approximate curvature using tensor-structured or diagonal estimates~\citep{gupta2018shampoo,anil2021scalable,liu2024sophia}. The open problem is how to obtain useful curvature information cheaply enough to improve end-to-end training.

Future work should explore:
\begin{enumerate}[leftmargin=*]
    \item adaptive curvature update schedules;
    \item selective curvature estimation for only some layers;
    \item low-precision curvature statistics;
    \item sharding-friendly preconditioners;
    \item curvature estimates that reuse information already computed during backpropagation;
    \item combinations of curvature estimation with low-rank or matrix-based updates.
\end{enumerate}

The practical test is whether curvature-aware methods improve wall-clock time or final performance under fixed compute, not merely whether they reduce the number of optimization steps.

\subsection{Optimizer Design for Long-Context Training}
\label{subsec:open_long_context}

Long-context LLM training changes the optimization and systems environment. Longer sequences increase activation memory, alter token correlations within a batch, and may require different learning-rate schedules or gradient accumulation strategies. Optimizer memory competes directly with activation memory, making memory-efficient optimizers more valuable.

Open questions include:
\begin{enumerate}[leftmargin=*]
    \item Do optimizers tuned for short-context pretraining remain optimal for long-context training?
    \item Does gradient noise behave differently as sequence length increases?
    \item Can memory-efficient optimizers enable longer contexts without sacrificing convergence?
    \item Should attention and feed-forward parameters use different update rules during long-context adaptation?
    \item How do optimizers interact with activation checkpointing and sequence parallelism?
\end{enumerate}

As context windows grow, optimizer design may become increasingly tied to memory management and sequence-level training dynamics.

\subsection{Fine-Tuning Versus Pretraining}
\label{subsec:open_finetuning_pretraining}

Pretraining and fine-tuning impose different optimizer requirements. Pretraining emphasizes long-horizon stability, throughput, and scaling behavior across massive token budgets. Fine-tuning often emphasizes memory feasibility, robustness to small datasets, and avoiding catastrophic forgetting.

An optimizer that is excellent for pretraining may not be ideal for full-parameter fine-tuning, and vice versa. \lomo{}, for example, is motivated by full-parameter fine-tuning under limited memory rather than large-scale pretraining~\citep{lv2024lomo}. Low-rank and memory-efficient optimizers may also have different value depending on whether the goal is training from scratch or adapting a pretrained model.

Future work should avoid treating LLM optimization as a single regime. Instead, optimizer recommendations should distinguish at least:
\begin{enumerate}[leftmargin=*]
    \item pretraining from scratch;
    \item continued pretraining;
    \item long-context adaptation;
    \item supervised fine-tuning;
    \item preference optimization and RL-style post-training;
    \item domain adaptation under limited hardware.
\end{enumerate}

Each regime may have different optimal trade-offs among memory, stability, and convergence.

\subsection{Optimizer Interaction with Post-Training}
\label{subsec:open_post_training}

Most optimizer discussions focus on next-token pretraining or supervised fine-tuning, but modern LLM development includes preference optimization, reinforcement learning, and other post-training stages. These stages often use smaller datasets, different losses, and more fragile training dynamics than pretraining.

Open questions include:
\begin{enumerate}[leftmargin=*]
    \item Are AdamW-style optimizers still optimal for preference optimization and RL fine-tuning?
    \item Do memory-efficient optimizers behave differently when gradients are high variance or reward-derived?
    \item Should post-training use different optimizer states from pretraining?
    \item Can curvature-aware or trust-region-like optimizers improve stability in alignment training?
    \item How do optimizer choices affect reward hacking, overoptimization, or loss of pretrained capabilities?
\end{enumerate}

This is an important direction because post-training often determines the behavior of deployed LLMs, yet optimizer choices in this stage are less systematically studied than pretraining optimizers.

\subsection{Hardware--Optimizer Co-Design}
\label{subsec:open_hardware_codesign}

Optimizer performance depends strongly on hardware. A method that is mathematically attractive may be impractical if it requires operations that are inefficient on accelerators. Conversely, an optimizer that maps well to matrix-multiply units or low-precision hardware may be highly effective in practice.

Future optimizer research should consider hardware co-design. Important questions include:
\begin{enumerate}[leftmargin=*]
    \item Which optimizer operations map efficiently to GPUs, TPUs, or future AI accelerators?
    \item Can matrix-based optimizers exploit high-throughput matrix multiplication units?
    \item Can optimizer states be stored or updated efficiently in low precision?
    \item How should optimizers be designed for memory-bandwidth-limited training?
    \item Can sharding, checkpointing, and optimizer updates be co-designed rather than layered independently?
\end{enumerate}

This perspective is especially relevant for matrix-based methods such as \muon{}, low-bit optimizers, and structured preconditioners. Their practical value depends on whether the hardware can execute their additional operations efficiently.

\subsection{Toward Adaptive and Learned Optimizer Selection}
\label{subsec:open_learned_optimizer_selection}

Most training recipes choose a single optimizer before training begins. However, the best optimizer may change over time. Early training may require stability and adaptivity, while late training may benefit from different regularization or lower-noise updates. Different layers may also benefit from different update rules.

A future direction is adaptive optimizer selection. Instead of using one optimizer globally, the training system could select update rules based on gradient statistics, layer type, training phase, or memory pressure. This could include switching from \adamw{} to a lower-memory method after a warmup phase, using curvature-aware updates only intermittently, or allocating richer optimizer states to layers with high sensitivity.

The challenge is to make such adaptation reliable. Learned or adaptive optimizer selection introduces additional complexity and risk. It also requires evaluation protocols that distinguish genuine improvements from overfitting to benchmark settings.

\subsection{Summary}
\label{subsec:open_summary}

The next generation of LLM optimizers will likely be shaped by several intertwined goals: reducing memory, improving update geometry, exploiting matrix and low-rank structure, scaling reliably, and integrating with distributed hardware. The most important open problems are not only algorithmic but also methodological and systems-oriented. The field needs fair benchmarks, optimizer scaling laws, stronger theory for matrix-based methods, and practical guidance for choosing optimizers under fixed resource budgets. These challenges motivate the concluding perspective of this survey: LLM optimizer research should be evaluated as a joint problem in optimization, architecture, and systems design.

\section{Conclusion}
\label{sec:conclusion}

Optimizers are a central component of large language model training. They determine not only how quickly a model descends the training loss, but also which model sizes, context lengths, batch sizes, and fine-tuning regimes are feasible under a fixed hardware budget. For this reason, optimizer design for LLMs cannot be evaluated purely as an abstract mathematical update rule. It must be understood as a joint problem in optimization, architecture, memory, numerics, and distributed systems.

This survey organized the optimizer landscape for LLMs around the metaphor of \emph{LLM Valley}: a rugged optimization landscape in which different methods navigate different trade-offs. \adamw{} remains the main trail through this valley. It is robust, well understood, widely implemented, and difficult to beat under strong tuning. Its combination of momentum, coordinate-wise adaptivity, and decoupled weight decay has made it the reference optimizer for Transformer-based language models. At the same time, its full first- and second-moment states impose substantial memory costs, and its coordinate-wise update geometry does not explicitly exploit the matrix structure of Transformer parameters.

The recent literature can be viewed as a set of attempts to move beyond this baseline along several directions. Memory-efficient optimizers such as \adafactor{}, 8-bit optimizers, Adam-mini, \lomo{}, \galore{}, and APOLLO-like methods aim to reduce the optimizer-state burden that makes full-parameter training expensive. Sign-based and discovered optimizers such as signSGD and \lion{} challenge the assumption that full gradient magnitudes and second-moment estimates are always necessary. Curvature-aware methods such as Shampoo and \sophia{} revisit second-order information in scalable forms. Low-rank and projection-based methods exploit structure in gradients and updates. Matrix-based optimizers such as \muon{} treat Transformer weights as matrices rather than independent coordinates, opening a new direction based on update geometry and orthogonalization.

A recurring theme is that no optimizer is universally best. The right choice depends on the training regime. Large-scale pretraining rewards long-horizon stability, throughput, and strong scaling. Full-parameter fine-tuning under limited memory rewards reduced optimizer state and low activation overhead. Long-context training places optimizer memory in direct competition with activation memory. Curvature-aware methods may be attractive when token efficiency dominates, while memory-efficient methods may be preferable when model size or sequence length is memory-bound. Matrix-based methods may offer benefits when update geometry matters, but require careful implementation and theoretical understanding.

This survey also emphasized that optimizer evaluation must be multi-objective. Validation loss as a function of tokens is important, but insufficient. Practical comparisons must also account for wall-clock time, peak memory, total optimizer-state size, checkpoint size, communication overhead, implementation quality, numerical precision, tuning effort, and downstream performance. In particular, comparisons against \adamw{} should use strong baselines and comparable hyperparameter search budgets. Otherwise, reported gains may reflect under-tuned baselines, small-scale artifacts, or implementation differences rather than genuine optimizer improvements.

Looking forward, several directions appear especially important. First, the field needs optimizer scaling laws that predict how methods behave as model size, batch size, data size, and context length increase. Second, more standardized benchmarks are needed to compare optimizers under fixed-model and fixed-resource regimes. Third, memory should be treated as an allocatable optimization resource: different parameters may warrant different levels of optimizer state. Fourth, matrix-based and orthogonalized optimizers require stronger theory connecting update geometry to Transformer training dynamics. Finally, optimizer research should be increasingly co-designed with hardware and distributed systems, since practical performance depends on the operations that accelerators can execute efficiently.

The emerging picture is not that \adamw{} will be replaced by a single universal successor. Rather, the optimizer landscape is becoming more specialized and architecture-aware. Future LLM training systems may combine multiple update rules: adaptive methods for some parameters, matrix-based updates for large projections, low-rank or memory-efficient updates where state is costly, and curvature-aware updates when better conditioning is worth the overhead. Navigating LLM Valley will therefore require not just new optimizers, but a deeper understanding of when, where, and why each optimizer works.

In summary, optimizers for LLMs are evolving from generic first-order update rules into system-aware, structure-aware training algorithms. The most promising methods will be those that improve the full training frontier: better loss, lower memory, faster wall-clock convergence, greater stability, and compatibility with the distributed infrastructure required for modern language models. A mature science of LLM optimization will need to integrate empirical benchmarking, theoretical analysis, and systems design into a unified framework.

\bibliography{refs}

@inproceedings{kingma2015adam,
  title     = {Adam: A Method for Stochastic Optimization},
  author    = {Kingma, Diederik P. and Ba, Jimmy},
  booktitle = {International Conference on Learning Representations},
  year      = {2015},
  url       = {https://arxiv.org/abs/1412.6980}
}

@article{duchi2011adagrad,
  title   = {Adaptive Subgradient Methods for Online Learning and Stochastic Optimization},
  author  = {Duchi, John and Hazan, Elad and Singer, Yoram},
  journal = {Journal of Machine Learning Research},
  volume  = {12},
  pages   = {2121--2159},
  year    = {2011},
  url     = {https://jmlr.org/papers/v12/duchi11a.html}
}

@misc{tieleman2012rmsprop,
  title        = {Lecture 6.5---RMSProp: Divide the Gradient by a Running Average of Its Recent Magnitude},
  author       = {Tieleman, Tijmen and Hinton, Geoffrey},
  howpublished = {COURSERA: Neural Networks for Machine Learning},
  year         = {2012}
}

@inproceedings{reddi2018adam,
  title     = {On the Convergence of Adam and Beyond},
  author    = {Reddi, Sashank J. and Kale, Satyen and Kumar, Sanjiv},
  booktitle = {International Conference on Learning Representations},
  year      = {2018},
  url       = {https://openreview.net/forum?id=ryQu7f-RZ}
}

@inproceedings{loshchilov2019decoupled,
  title     = {Decoupled Weight Decay Regularization},
  author    = {Loshchilov, Ilya and Hutter, Frank},
  booktitle = {International Conference on Learning Representations},
  year      = {2019},
  url       = {https://openreview.net/forum?id=Bkg6RiCqY7}
}

@article{ruder2016overview,
  title   = {An Overview of Gradient Descent Optimization Algorithms},
  author  = {Ruder, Sebastian},
  journal = {arXiv preprint arXiv:1609.04747},
  year    = {2016},
  url     = {https://arxiv.org/abs/1609.04747}
}

@article{bottou2018optimization,
  title   = {Optimization Methods for Large-Scale Machine Learning},
  author  = {Bottou, L{\'e}on and Curtis, Frank E. and Nocedal, Jorge},
  journal = {SIAM Review},
  volume  = {60},
  number  = {2},
  pages   = {223--311},
  year    = {2018},
  doi     = {10.1137/16M1080173},
  url     = {https://doi.org/10.1137/16M1080173}
}

@article{sun2019survey,
  title   = {A Survey of Optimization Methods from a Machine Learning Perspective},
  author  = {Sun, Shiliang and Cao, Zehui and Zhu, Han and Zhao, Jing},
  journal = {IEEE Transactions on Cybernetics},
  volume  = {50},
  number  = {8},
  pages   = {3668--3681},
  year    = {2020},
  doi     = {10.1109/TCYB.2019.2950779},
  url     = {https://arxiv.org/abs/1906.06821}
}

@inproceedings{vaswani2017attention,
  title     = {Attention Is All You Need},
  author    = {Vaswani, Ashish and Shazeer, Noam and Parmar, Niki and Uszkoreit, Jakob and Jones, Llion and Gomez, Aidan N. and Kaiser, Lukasz and Polosukhin, Illia},
  booktitle = {Advances in Neural Information Processing Systems},
  year      = {2017},
  url       = {https://arxiv.org/abs/1706.03762}
}

@inproceedings{devlin2019bert,
  title     = {{BERT}: Pre-training of Deep Bidirectional Transformers for Language Understanding},
  author    = {Devlin, Jacob and Chang, Ming-Wei and Lee, Kenton and Toutanova, Kristina},
  booktitle = {Proceedings of the 2019 Conference of the North American Chapter of the Association for Computational Linguistics},
  pages     = {4171--4186},
  year      = {2019},
  url       = {https://aclanthology.org/N19-1423/}
}

@inproceedings{brown2020language,
  title     = {Language Models are Few-Shot Learners},
  author    = {Brown, Tom B. and Mann, Benjamin and Ryder, Nick and Subbiah, Melanie and Kaplan, Jared and Dhariwal, Prafulla and Neelakantan, Arvind and Shyam, Pranav and Sastry, Girish and Askell, Amanda and others},
  booktitle = {Advances in Neural Information Processing Systems},
  volume    = {33},
  pages     = {1877--1901},
  year      = {2020},
  url       = {https://arxiv.org/abs/2005.14165}
}

@article{chowdhery2023palm,
  title   = {{PaLM}: Scaling Language Modeling with Pathways},
  author  = {Chowdhery, Aakanksha and Narang, Sharan and Devlin, Jacob and Bosma, Maarten and Mishra, Gaurav and Roberts, Adam and Barham, Paul and Chung, Hyung Won and Sutton, Charles and Gehrmann, Sebastian and others},
  journal = {Journal of Machine Learning Research},
  volume  = {24},
  number  = {240},
  pages   = {1--113},
  year    = {2023},
  url     = {https://jmlr.org/papers/v24/22-1144.html}
}

@article{touvron2023llama,
  title   = {{LLaMA}: Open and Efficient Foundation Language Models},
  author  = {Touvron, Hugo and Lavril, Thibaut and Izacard, Gautier and Martinet, Xavier and Lachaux, Marie-Anne and Lacroix, Timoth{\'e}e and Rozi{\`e}re, Baptiste and Goyal, Naman and Hambro, Eric and Azhar, Faisal and others},
  journal = {arXiv preprint arXiv:2302.13971},
  year    = {2023},
  url     = {https://arxiv.org/abs/2302.13971}
}

@article{touvron2023llama2,
  title   = {{Llama 2}: Open Foundation and Fine-Tuned Chat Models},
  author  = {Touvron, Hugo and Martin, Louis and Stone, Kevin and Albert, Peter and Almahairi, Amjad and Babaei, Yasmine and Bashlykov, Nikolay and Batra, Soumya and Bhargava, Prajjwal and Bhosale, Shruti and others},
  journal = {arXiv preprint arXiv:2307.09288},
  year    = {2023},
  url     = {https://arxiv.org/abs/2307.09288}
}

@article{dubey2024llama3,
  title   = {The {Llama 3} Herd of Models},
  author  = {Dubey, Abhimanyu and Jauhri, Abhinav and Pandey, Abhinav and Kadian, Abhishek and Al-Dahle, Ahmad and Letman, Aiesha and Mathur, Akhil and Schelten, Alan and Yang, Amy and Fan, Angela and others},
  journal = {arXiv preprint arXiv:2407.21783},
  year    = {2024},
  url     = {https://arxiv.org/abs/2407.21783}
}

@article{jiang2023mistral,
  title   = {{Mistral 7B}},
  author  = {Jiang, Albert Q. and Sablayrolles, Alexandre and Mensch, Arthur and Bamford, Chris and Chaplot, Devendra Singh and de las Casas, Diego and Bressand, Florian and Lengyel, Gianna and Lample, Guillaume and Saulnier, Lucile and others},
  journal = {arXiv preprint arXiv:2310.06825},
  year    = {2023},
  url     = {https://arxiv.org/abs/2310.06825}
}

@article{tay2023efficient,
  title   = {Efficient Transformers: A Survey},
  author  = {Tay, Yi and Dehghani, Mostafa and Bahri, Dara and Metzler, Donald},
  journal = {ACM Computing Surveys},
  volume  = {55},
  number  = {6},
  pages   = {1--28},
  year    = {2023},
  doi     = {10.1145/3530811}
}

@article{wan2023efficienttraining,
  title   = {A Survey on Efficient Training of Transformers},
  author  = {Wan, Zhongwei and Wang, Xin and Liu, Che and Alam, Samiul and Zheng, Yu and Liu, Jiachen and Qu, Zhuoran and Yan, Shen and Zhu, Yi and Zhang, Quanlu and others},
  journal = {arXiv preprint arXiv:2302.01107},
  year    = {2023},
  url     = {https://arxiv.org/abs/2302.01107}
}

@inproceedings{shazeer2018adafactor,
  title     = {Adafactor: Adaptive Learning Rates with Sublinear Memory Cost},
  author    = {Shazeer, Noam and Stern, Mitchell},
  booktitle = {Proceedings of the 35th International Conference on Machine Learning},
  series    = {Proceedings of Machine Learning Research},
  volume    = {80},
  pages     = {4596--4604},
  publisher = {PMLR},
  year      = {2018},
  url       = {https://proceedings.mlr.press/v80/shazeer18a.html}
}

@inproceedings{you2020large,
  title     = {Large Batch Optimization for Deep Learning: Training {BERT} in 76 Minutes},
  author    = {You, Yang and Li, Jing and Reddi, Sashank and Hseu, Jonathan and Kumar, Sanjiv and Bhojanapalli, Srinadh and Song, Xiaodan and Demmel, James and Keutzer, Kurt and Hsieh, Cho-Jui},
  booktitle = {International Conference on Learning Representations},
  year      = {2020},
  url       = {https://openreview.net/forum?id=Syx4wnEtvH}
}

@inproceedings{dettmers2022optimizers,
  title     = {8-bit Optimizers via Block-wise Quantization},
  author    = {Dettmers, Tim and Lewis, Mike and Belkada, Younes and Zettlemoyer, Luke},
  booktitle = {International Conference on Learning Representations},
  year      = {2022},
  url       = {https://openreview.net/forum?id=shpkpVXzo3h}
}

@inproceedings{zhang2025adammini,
  title     = {Adam-mini: Use Fewer Learning Rates To Gain More},
  author    = {Zhang, Yushun and Chen, Congliang and Li, Ziniu and Ding, Tian and Wu, Chenwei and Kingma, Diederik P. and Ye, Yinyu and Luo, Zhi-Quan and Sun, Ruoyu},
  booktitle = {International Conference on Learning Representations},
  year      = {2025},
  url       = {https://openreview.net/forum?id=iBExhaU3Lc}
}

@inproceedings{lv2024lomo,
  title     = {Full Parameter Fine-tuning for Large Language Models with Limited Resources},
  author    = {Lv, Kai and Yang, Yuqing and Liu, Tengxiao and Gao, Qinghui and Guo, Qipeng and Qiu, Xipeng},
  booktitle = {Proceedings of the 62nd Annual Meeting of the Association for Computational Linguistics},
  year      = {2024},
  url       = {https://aclanthology.org/2024.acl-long.445/}
}

@article{han2024apollo,
  title   = {{APOLLO}: {SGD}-like Memory, {AdamW}-level Performance},
  author  = {Zhu, Hanqing and Zhang, Zhenyu and Cong, Wenyan and Liu, Xi and Park, Sem and Chandra, Vikas and Long, Bo and Pan, David Z. and Wang, Zhangyang and Lee, Jinwon},
  journal = {arXiv preprint arXiv:2412.05270},
  year    = {2024},
  url     = {https://arxiv.org/abs/2412.05270}
}

@inproceedings{bernstein2018signsgd,
  title     = {signSGD: Compressed Optimisation for Non-Convex Problems},
  author    = {Bernstein, Jeremy and Wang, Yu-Xiang and Azizzadenesheli, Kamyar and Anandkumar, Anima},
  booktitle = {Proceedings of the 35th International Conference on Machine Learning},
  series    = {Proceedings of Machine Learning Research},
  volume    = {80},
  pages     = {560--569},
  publisher = {PMLR},
  year      = {2018},
  url       = {https://proceedings.mlr.press/v80/bernstein18a.html}
}

@inproceedings{chen2023symbolic,
  title     = {Symbolic Discovery of Optimization Algorithms},
  author    = {Chen, Xiangning and Liang, Chen and Huang, Da and Real, Esteban and Wang, Kaiyuan and Liu, Yao and Pham, Hieu and Dong, Xuanyi and Luong, Thang and Hsieh, Cho-Jui and Lu, Yifeng and Le, Quoc V.},
  booktitle = {Advances in Neural Information Processing Systems},
  year      = {2023},
  url       = {https://arxiv.org/abs/2302.06675}
}

@inproceedings{gupta2018shampoo,
  title     = {Shampoo: Preconditioned Stochastic Tensor Optimization},
  author    = {Gupta, Vineet and Koren, Tomer and Singer, Yoram},
  booktitle = {Proceedings of the 35th International Conference on Machine Learning},
  series    = {Proceedings of Machine Learning Research},
  volume    = {80},
  pages     = {1842--1850},
  publisher = {PMLR},
  year      = {2018},
  url       = {https://proceedings.mlr.press/v80/gupta18a.html}
}

@article{anil2021scalable,
  title   = {Scalable Second Order Optimization for Deep Learning},
  author  = {Anil, Rohan and Gupta, Vineet and Koren, Tomer and Regan, Kevin and Singer, Yoram},
  journal = {arXiv preprint arXiv:2002.09018},
  year    = {2021},
  url     = {https://arxiv.org/abs/2002.09018}
}

@inproceedings{liu2024sophia,
  title     = {Sophia: A Scalable Stochastic Second-order Optimizer for Language Model Pre-training},
  author    = {Liu, Hong and Li, Zhiyuan and Hall, David and Liang, Percy and Ma, Tengyu},
  booktitle = {International Conference on Learning Representations},
  year      = {2024},
  url       = {https://openreview.net/forum?id=3xHDeA8Noi}
}

@article{liu1989lbfgs,
  title   = {On the Limited Memory Method for Large Scale Optimization},
  author  = {Liu, Dong C. and Nocedal, Jorge},
  journal = {Mathematical Programming},
  volume  = {45},
  number  = {1},
  pages   = {503--528},
  year    = {1989},
  doi     = {10.1007/BF01589116}
}

@book{nocedal2006numerical,
  title     = {Numerical Optimization},
  author    = {Nocedal, Jorge and Wright, Stephen J.},
  edition   = {2},
  publisher = {Springer},
  year      = {2006},
  doi       = {10.1007/978-0-387-40065-5}
}

@inproceedings{schraudolph2007stochastic,
  title     = {A Stochastic Quasi-Newton Method for Online Convex Optimization},
  author    = {Schraudolph, Nicol N. and Yu, Jin and G{\"u}nter, Simon},
  booktitle = {Proceedings of the Eleventh International Conference on Artificial Intelligence and Statistics},
  year      = {2007},
  url       = {https://proceedings.mlr.press/v2/schraudolph07a.html}
}

@article{mokhtari2015global,
  title   = {Global Convergence of Online Limited Memory BFGS},
  author  = {Mokhtari, Aryan and Ribeiro, Alejandro},
  journal = {Journal of Machine Learning Research},
  volume  = {16},
  pages   = {3151--3181},
  year    = {2015},
  url     = {https://jmlr.org/papers/v16/mokhtari15a.html}
}

@article{goldfarb2020practical,
  title   = {Practical Quasi-Newton Methods for Training Deep Neural Networks},
  author  = {Goldfarb, Donald and Ren, Yi and Bahamou, Achraf},
  journal = {arXiv preprint arXiv:2006.08877},
  year    = {2020},
  url     = {https://arxiv.org/abs/2006.08877}
}

@article{niu2023mlbfgs,
  title   = {{mL-BFGS}: A Momentum-Based L-BFGS for Distributed Large-Scale Neural Network Optimization},
  author  = {Niu, Yue and Fabian, Zalan and Lee, Sunwoo and Soltanolkotabi, Mahdi and Avestimehr, Salman},
  journal = {arXiv preprint arXiv:2307.13744},
  year    = {2023},
  url     = {https://arxiv.org/abs/2307.13744}
}

@article{ranganath2025sr1cubic,
  title   = {Symmetric Rank-One Quasi-Newton Methods for Deep Learning Using Cubic Regularization},
  author  = {Ranganath, Aditya and Singhal, Mukesh and Marcia, Roummel},
  journal = {arXiv preprint arXiv:2502.12298},
  year    = {2025},
  url     = {https://arxiv.org/abs/2502.12298}
}

@inproceedings{zhao2024galore,
  title     = {{GaLore}: Memory-Efficient {LLM} Training by Gradient Low-Rank Projection},
  author    = {Zhao, Jiawei and Zhang, Zhenyu and Chen, Beidi and Wang, Zhangyang and Anandkumar, Anima and Tian, Yuandong},
  booktitle = {Proceedings of the 41st International Conference on Machine Learning},
  series    = {Proceedings of Machine Learning Research},
  publisher = {PMLR},
  year      = {2024},
  url       = {https://proceedings.mlr.press/v235/zhao24s.html}
}

@misc{jordan2024muon,
  title        = {{Muon}: An Optimizer for Hidden Layers in Neural Networks},
  author       = {Jordan, Keller},
  year         = {2024},
  howpublished = {Blog post and software repository},
  url          = {https://kellerjordan.github.io/posts/muon/},
  note         = {Introduces MomentUm Orthogonalized by Newton--Schulz.}
}

@article{liu2025muon,
  title   = {Muon is Scalable for {LLM} Training},
  author  = {Liu, Jingyuan and Su, Jianlin and Yao, Xingcheng and Jiang, Zhejun and Lai, Guokun and Du, Yulun and Qin, Yidao and Xu, Weixin and Lu, Enzhe and Yan, Junjie and Chen, Yanru and Zheng, Huabin and Liu, Yibo and Liu, Shaowei and Yin, Bohong and He, Weiran and Zhu, Han and Wang, Yuzhi and Wang, Jianzhou and Dong, Mengnan and Zhang, Zheng and Kang, Yongsheng and Zhang, Hao and Xu, Xinran and Zhang, Yutao and Wu, Yuxin and Zhou, Xinyu and Yang, Zhilin},
  journal = {arXiv preprint arXiv:2502.16982},
  year    = {2025},
  url     = {https://arxiv.org/abs/2502.16982}
}

@article{kim2026convergence,
  title   = {Convergence of {Muon} with Newton--Schulz},
  author  = {Kim, Gyu Yeol and Oh, Min-hwan},
  journal = {arXiv preprint arXiv:2601.19156},
  year    = {2026},
  url     = {https://arxiv.org/abs/2601.19156}
}

@inproceedings{zhao2025deconstructing,
  title     = {Deconstructing What Makes a Good Optimizer for Autoregressive Language Models},
  author    = {Zhao, Rosie and Morwani, Depen and Brandfonbrener, David and Vyas, Nikhil and Kakade, Sham},
  booktitle = {International Conference on Learning Representations},
  year      = {2025},
  url       = {https://openreview.net/forum?id=0J9gL2DVO4}
}

@article{semenov2025benchmarking,
  title   = {Benchmarking Optimizers for Large Language Model Pretraining},
  author  = {Semenov, Andrei and Pagliardini, Matteo and Jaggi, Martin},
  journal = {arXiv preprint arXiv:2509.01440},
  year    = {2025},
  url     = {https://arxiv.org/abs/2509.01440}
}

@inproceedings{zhang2024whytransformersneedadam,
  title     = {Why Transformers Need Adam: A Hessian Perspective},
  author    = {Zhang, Yushun and Chen, Congliang and Li, Ziniu and Ding, Tian and Wu, Chenwei and Ye, Yinyu and Luo, Zhi-Quan and Sun, Ruoyu},
  booktitle = {Advances in Neural Information Processing Systems},
  year      = {2024},
  url       = {https://proceedings.neurips.cc/paper_files/paper/2024/hash/ee0e45ff4de76cbfdf07015a7839f339-Abstract-Conference.html}
}

@article{tomihari2025gradientheterogeneity,
  title   = {Understanding Why Adam Outperforms SGD: Gradient Heterogeneity in Transformers},
  author  = {Tomihari, Akiyoshi and Sato, Issei},
  journal = {arXiv preprint arXiv:2502.00213},
  year    = {2025},
  url     = {https://arxiv.org/abs/2502.00213}
}

@article{molybog2023adaminstability,
  title   = {A Theory on Adam Instability in Large-Scale Machine Learning},
  author  = {Molybog, Igor and Albert, Peter and Chen, Moya and DeVito, Zachary and Esiobu, David and Goyal, Naman and Koura, Punit Singh and Narang, Sharan and Poulton, Andrew and Silva, Ruan and Tang, Binh and Liskovich, Diana and Xu, Puxin and Zhang, Yuchen and Kambadur, Melanie and Roller, Stephen and Zhang, Susan},
  journal = {arXiv preprint arXiv:2304.09871},
  year    = {2023},
  url     = {https://arxiv.org/abs/2304.09871}
}

@inproceedings{micikevicius2018mixed,
  title     = {Mixed Precision Training},
  author    = {Micikevicius, Paulius and Narang, Sharan and Alben, Jonah and Diamos, Gregory and Elsen, Erich and Garcia, David and Ginsburg, Boris and Houston, Michael and Kuchaiev, Oleksii and Venkatesh, Ganesh and Wu, Hao},
  booktitle = {International Conference on Learning Representations},
  year      = {2018},
  url       = {https://openreview.net/forum?id=r1gs9JgRZ}
}

@article{zhao2019adaptive,
  title   = {Adaptive Loss Scaling for Mixed Precision Training},
  author  = {Zhao, Ruizhe and Vogel, Brian and Ahmed, Tanvir},
  journal = {arXiv preprint arXiv:1910.12385},
  year    = {2019},
  url     = {https://arxiv.org/abs/1910.12385}
}

@article{wang2023stableadam16bit,
  title   = {Stable Adam Optimization for 16-bit Neural Networks Training},
  author  = {Wang, Yifei and others},
  journal = {arXiv preprint arXiv:2307.16189},
  year    = {2023},
  url     = {https://arxiv.org/abs/2307.16189},
  note    = {Author metadata should be verified before final submission.}
}

@article{sergeev2018horovod,
  title   = {Horovod: Fast and Easy Distributed Deep Learning in TensorFlow},
  author  = {Sergeev, Alexander and Del Balso, Mike},
  journal = {arXiv preprint arXiv:1802.05799},
  year    = {2018},
  url     = {https://arxiv.org/abs/1802.05799}
}

@misc{pytorchddp,
  title        = {{PyTorch DistributedDataParallel} Design Note},
  author       = {{PyTorch Contributors}},
  howpublished = {\url{https://docs.pytorch.org/docs/stable/notes/ddp.html}},
  year         = {2024},
  note         = {Accessed 2026-05-09}
}

@article{shoeybi2019megatron,
  title   = {{Megatron-LM}: Training Multi-Billion Parameter Language Models Using Model Parallelism},
  author  = {Shoeybi, Mohammad and Patwary, Mostofa and Puri, Raul and LeGresley, Patrick and Casper, Jared and Catanzaro, Bryan},
  journal = {arXiv preprint arXiv:1909.08053},
  year    = {2019},
  url     = {https://arxiv.org/abs/1909.08053}
}

@inproceedings{rajbhandari2020zero,
  title     = {{ZeRO}: Memory Optimizations Toward Training Trillion Parameter Models},
  author    = {Rajbhandari, Samyam and Rasley, Jeff and Ruwase, Olatunji and He, Yuxiong},
  booktitle = {Proceedings of the International Conference for High Performance Computing, Networking, Storage and Analysis},
  year      = {2020},
  url       = {https://arxiv.org/abs/1910.02054}
}

@inproceedings{narayanan2021efficient,
  title     = {Efficient Large-Scale Language Model Training on {GPU} Clusters Using {Megatron-LM}},
  author    = {Narayanan, Deepak and Shoeybi, Mohammad and Casper, Jared and LeGresley, Patrick and Patwary, Mostofa and Korthikanti, Vijay and Vainbrand, Dmitri and Kashinkunti, Prethvi and Bernauer, Julie and Catanzaro, Bryan and Phanishayee, Amar and Zaharia, Matei},
  booktitle = {Proceedings of the International Conference for High Performance Computing, Networking, Storage and Analysis},
  year      = {2021},
  url       = {https://arxiv.org/abs/2104.04473}
}

@article{zhao2023fsdp,
  title   = {{PyTorch FSDP}: Experiences on Scaling Fully Sharded Data Parallel},
  author  = {Zhao, Yanli and Gu, Andrew and Varma, Rohan and Luo, Liang and Huang, Chien-Chin and Xu, Min and Wright, Less and Shojanazeri, Hamid and Ott, Myle and Shleifer, Sam and Desmaison, Alban and Balioglu, Can and Nguyen, Bernard and Chauhan, Geeta and Hao, Yuchen and Li, Shen},
  journal = {Proceedings of the VLDB Endowment},
  volume  = {16},
  number  = {12},
  pages   = {3848--3860},
  year    = {2023},
  url     = {https://arxiv.org/abs/2304.11277}
}

\end{document}